%% file: main.tex
\documentclass[table]{gtech}
\PassOptionsToPackage{table, usenames, dvipsnames}{xcolor}
\usepackage{bigdelim}
\usepackage{longtable}
\usepackage{tabularray}
\usepackage{float}
\usepackage{datatool}
\usepackage[justification=centering]{caption}

\RequirePackage{tgpagella} 
\RequirePackage{mathpazo}  
\usepackage{times}
\usepackage{latexsym}
\usepackage[T1]{fontenc}
\usepackage[utf8]{inputenc}
\usepackage{microtype}

\newcommand{\ignore}[1]{}

\renewcommand{\title}[1]{\newcommand{\titlelist}{{\huge\selectfont #1}}}

\usepackage{arydshln}
\definecolor{CQColor}{rgb}{0.0,0.0,1.0} 

\usepackage{pifont}
\usepackage{tabulary}
\usepackage{fontawesome5}
\usepackage{bbding}
\usepackage{multicol}

\newlength\savewidth

\usepackage{hyperref}
\usepackage{url}

\usepackage{makecell}  
\usepackage{graphicx}
\usepackage{multirow}
\usepackage{color}
\usepackage{array}
\usepackage{algorithm}
\usepackage{algpseudocode}
\usepackage{wrapfig}
\usepackage{amsmath}
\usepackage{makecell}
\usepackage{colortbl}
\usepackage{xcolor}
\usepackage{caption}
\usepackage{subcaption}  
\definecolor{darkred}{RGB}{162, 0, 0}

\definecolor{darkblue}{RGB}{4, 6, 173}

\definecolor{darkgreen}{RGB}{61, 134, 73}

\algnewcommand{\algorithmicinit}{\textbf{Initialize:}}
\algnewcommand{\Init}{\algorithmicinit}
\usepackage{amsmath,amsfonts,bm}
\usepackage{amsthm}
\usepackage{xspace}
\usepackage[most]{tcolorbox}
\usepackage{enumitem}
\usepackage{amssymb}
\usepackage{mathtools}
\usepackage{physics} 
\usepackage{thmtools}

\usepackage{amsfonts,bm}
\usepackage{dsfont}  
\usepackage{diagbox} 

\usepackage{tabularx}
\tcbuselibrary{skins}

\usepackage{titletoc}   

\usepackage{booktabs}

\tcbuselibrary{breakable}
\definecolor{my_green}{RGB}{51,102,0}
\definecolor{my_purple}{RGB}{160, 43, 147}
\definecolor{my_blue}{RGB}{15, 158, 213}

\NewDocumentCommand{\ganqu}
{ mO{} }{\textcolor{blue}{\textsuperscript{\textit{ganqu}}\textsf{\textbf{\small[#1]}}}}
\NewDocumentCommand{\yafu}
{ mO{} }{\textcolor{cyan}{\textsuperscript{\textit{yafu}}\textsf{\textbf{\small[#1]}}}}

\NewDocumentCommand{\jianhao}
{ mO{} }{\textcolor{red}{\textsuperscript{\textit{jianhao}}\textsf{\textbf{\small[#1]}}}}

\definecolor{deltaBg}{RGB}{220,230,255} 

\newtcolorbox{remark}[1][]{enhanced,
  breakable,
  colback=violet!6,           
  colframe=violet!50!black,   
  coltitle=black,             
  colbacktitle=violet!18,     
  fonttitle=\bfseries,
  title=Remark,
  #1}

\definecolor{uclablue}{rgb}{0.15, 0.45, 0.68}
\hypersetup{
    breaklinks,
    citecolor=uclablue,
    colorlinks=true,
}

\definecolor{lightgreen}{RGB}{0,150,0}
\definecolor{myred}{RGB}{200,0,0} 

\title{\textbf{DE-Venus: A Data-Efficient RLVR Framework for Large Language Models}}
\author[1,2*]{Shenzhi Yang}
\author[1,2*]{Guangcheng Zhu}
\author[1,2*]{Kai Tang}
\author[1,2*]{Zhengqing Zang}
\author[2]{Xing Zheng}
\author[1\ddag]{Haobo Wang}
\author[2]{Yingfan Ma}
\author[2\ddag]{Bowen Song}
\author[3]{Bo Han}
\author[4]{Bo An}
\author[5]{Lei Feng}
\author[2]{Weiqiang Wang}
\author[1]{Junbo Zhao}
\author[1]{Gang Chen}

\affiliation[1]{Zhejiang University \,\, $^2$Ant Group \,\, $^3$Hong Kong Baptist University  \\ $^4$Nanyang Technological University \,\, $^5$Southeast University \\[0.5em]}

\contribution[*]{Equal contribution}
\contribution[\ddag]{Corresponding authors}

\abstract{\fontsize{11pt}{12pt} \textit{Reinforcement learning with verifiable rewards (RLVR) improves large language model reasoning, but its practical scaling is constrained by expensive on-policy rollouts and the cost of obtaining reliable targets at scale. Existing methods address sample selection, incomplete supervision, or noisy labels separately, often entangling supervision logic with distributed training and hindering controlled comparison and reuse. We present \textbf{DE-Venus}, a unified framework for data-efficient RLVR that treats supervision as evolving state across data preparation and policy optimization. It organizes this lifecycle into three modules: Active Data Selection allocates training and annotation budgets; Weak Supervision Construction derives learning signals from unlabeled examples; and Training-Time Supervision Refinement filters or corrects unreliable supervision. DE-Venus supports seven representative methods and a data-selection pipeline by expressing method-specific decisions as offline dataset transitions or online transformations of targets, rewards, batches, and advantages while preserving \textit{verl}'s distributed execution contracts. Across public benchmarks and three business scenarios, separate configurations preserve or improve model quality with only 10\% of labels or as little as 13\% of relevant data; selected business configurations also reduce observed convergence steps by 63\%--75\%. DE-Venus thus reduces annotation and training costs without sacrificing scalable RL execution.}}

\addsecondlogo[0.95cm]{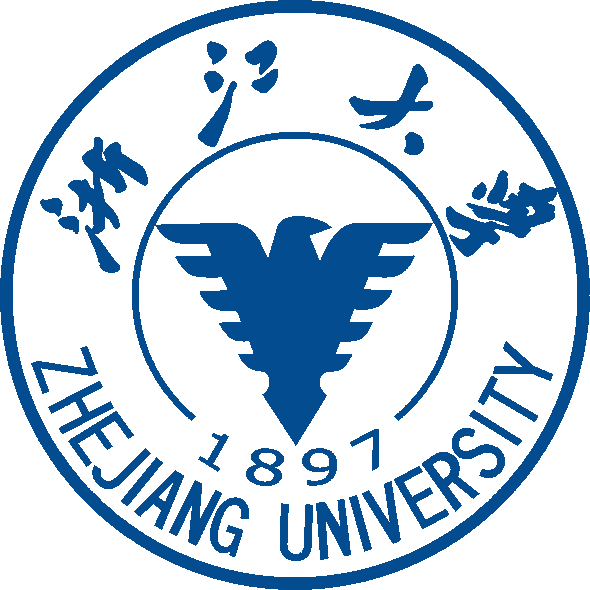}
\begin{document}
\maketitle

\input{sections/intro}
\input{sections/devenus_design}
\input{sections/method}
\input{sections/derl_experiments}

\input{sections/conclusion}

\bibliographystyle{assets/plainnat}
\bibliography{main}

\newpage
\appendix

\end{document}

%% file: sections/intro.tex
\section{Introduction}

Reinforcement learning with verifiable rewards (RLVR) has become a prominent paradigm for improving the reasoning capabilities of large language models (LLMs)~\citep{jaech2024openai,guo2025deepseek,team2025kimi,yang2025qwen3}. Given a question, the current policy samples multiple reasoning trajectories, an outcome verifier evaluates their final answers, and a policy optimizer such as Group Relative Policy Optimization (GRPO) reinforces responses that outperform others in the group~\citep{grpo,drgrpo,yu2025dapo,zheng2025group}. This loop has produced substantial gains on mathematics, code, and related reasoning tasks~\citep{guo2025deepseek,team2025kimi,orz,li2025limr}. Its effectiveness, however, assumes that every training question merits repeated rollout computation and is paired with a reliable, readily verifiable target.

These assumptions are difficult to sustain at scale. On-policy RLVR repeatedly incurs generation and optimization costs, even for questions that are mastered, intractable, or unlikely to yield informative reward variation. Recent studies show that controlling sample difficulty and utility can materially affect optimization efficiency~\citep{yu2025dapo,bae2025online,zeng2025cures,li2025limr}. Reference answers are labor-intensive to construct, and reference-based verifiers can themselves be imperfect~\citep{yan2025verifybench}; the bottleneck is sharper in specialized domains that require scarce expertise. Data efficiency in RLVR is therefore not simply a matter of reducing the training-set size. It concerns the entire supervision lifecycle: \emph{which} questions should enter training, \emph{which} warrant external annotation, \emph{how} unlabeled questions can provide useful learning signals, and \emph{when} unreliable supervision should be filtered or repaired.

Prior work addresses individual stages of this lifecycle. Data-selection methods identify informative examples using verified difficulty, uncertainty, or learning dynamics~\citep{yu2025dapo,bae2025online,zeng2025cures,zhu2026pivottrace}. Unsupervised and label-free methods replace external targets with signals derived from rollout agreement, entropy, self-certainty, or self-play~\citep{zuo2025ttrl,agarwal2025unreasonable,zhao2025learning,zhao2025absolutezero}. Semi-supervised methods use a small trusted subset to assess supervision constructed over a larger unlabeled pool~\citep{yang2025trapo,zhu2026geomin}, while noisy-label methods use policy-generated evidence to diagnose or revise questionable targets~\citep{yang2026can}. Although these directions are complementary, their implementations remain fragmented. New methods often arrive as standalone pipelines or trainer forks that rewrite the rollout-to-update loop. Differences in data conventions, reward placement, filtering, state management, and execution then become entangled with the algorithm, complicating faithful reproduction, controlled comparison, and transfer across supervision settings.

We present \textbf{DE-Venus}, a unified framework for \textbf{d}ata-\textbf{e}fficient reinforcement learning for LLM reasoning. Its central abstraction treats supervision as evolving state whose source, reliability, and persistence may change from data preparation to policy optimization. Accordingly, DE-Venus separates \emph{how supervision is selected, constructed, and revised} from \emph{how distributed RL is executed}. It retains \textit{verl} as the execution substrate for rollout generation, optimization, validation, and checkpointing~\citep{sheng2024hybridflow_verl}, while localizing method-specific decisions to the points at which supervision semantics change: before training through data selection, during training through supervision construction or reliability-aware updates, and between rounds through persistent dataset transitions. This contract-preserving, ``minimal-invasion'' design allows heterogeneous methods to share one scalable backend without flattening their distinct learning objectives.

DE-Venus organizes the supervision lifecycle into three complementary modules. \textbf{Active Data Selection} allocates training and annotation budgets by determining which examples should be retained, externally labeled, or routed to weak supervision; its implementations cover verified-difficulty filtering, model-derived uncertainty, and probe-calibrated selection~\citep{yu2025dapo,bae2025online,zeng2025cures,zhu2026pivottrace}. \textbf{Weak Supervision Construction} enables unlabeled examples to participate in RLVR by inferring pseudo targets, enforcing cross-view agreement, or assigning target-free rewards, covering TTRL, Co-Rewarding, EMRL, and Intuitor~\citep{zuo2025ttrl,zhang2025corewarding,agarwal2025unreasonable,zhao2025learning}. \textbf{Training-Time Supervision Refinement} reassesses constructed or existing supervision using reward dynamics, representation geometry, and rollout evidence, supporting TraPO, GeoMin, and online label refinement~\citep{yang2025trapo,zhu2026geomin,yang2026can}. These modules identify distinct intervention points rather than mandatory consecutive stages, so each experiment activates only the mechanisms required by its supervision setting.

The framework provides common abstractions at the data and execution boundaries. A shared Parquet convention records persistent identity and supervision state, while verl-compatible batches carry targets, rewards, aligned selections, and advantages through online optimization. Reusable services provide voting, verification, uncertainty estimation, reward construction, evidence analysis, and validation logging. Configuration-driven dispatch exposes method-specific choices without introducing a second backend, allowing new supervision rules to be localized instead of rebuilding distributed rollout, optimization, and evaluation.

Our contributions are summarized as follows:
\begin{itemize}[leftmargin=1.5em]
    \item We formulate data-efficient RLVR as a unified supervision lifecycle spanning active data selection, weak supervision construction, and training-time supervision refinement.
    \item We develop DE-Venus, a contract-preserving framework over verl that localizes supervision-specific interventions while retaining a common scalable execution backend.
    \item We provide reproducible implementations and workflows for seven representative methods and an active data-selection pipeline, evaluated on public reasoning benchmarks and three business scenarios.
\end{itemize}

On standard mathematical and general-reasoning benchmarks, DE-Venus configurations surpass fully supervised references using only 10\% of the labels or 57.9\% of the training data with 29.3\% annotations. More importantly, the business scenarios demonstrate operational gains beyond controlled benchmarks. With a fixed 500-label budget, reliability-aware weak supervision improves the normalized credit-assignment metric by up to 14 points. Trajectory-based filtering removes 28\% of medical-empathy training examples while remaining within 1.9 points of full-data training. For intrinsic-safety training, retaining only 13\%--30.7\% of relevant source data preserves core safety metrics, improves selected capabilities by up to 6.7\%, and reduces observed convergence steps by 63\%--75\%. These results show that DE-Venus can reduce annotation requirements, training volume, and iteration time while preserving or improving model quality in practical \footnote{The code is available via \url{https://github.com/ZJU-DIG/DE-Venus}}.

%% file: sections/devenus_design.tex
\begin{figure*}[t]
\centering
\includegraphics[width=0.99\textwidth]{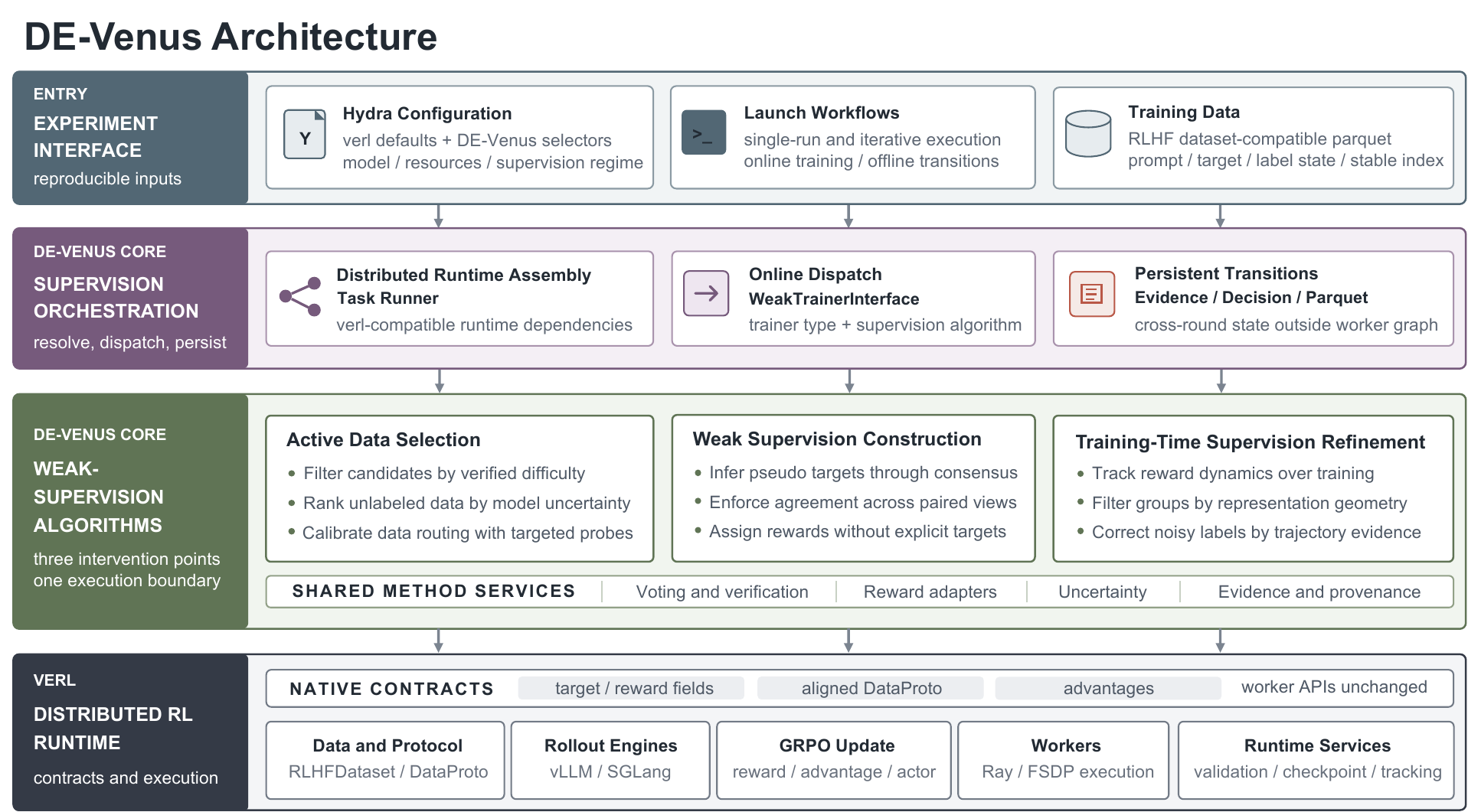}
\caption{\small DE-Venus architecture. Three supervision modules share a common orchestration layer and interact with verl through native contracts. The dashed path denotes persistent Parquet transitions across runs.}
\label{fig:devenus-architecture}
\end{figure*}

\section{Design and Implementation}
\label{sec:devenus-design}
DE-Venus provides a unified system for data-efficient RLVR when supervision is incomplete, unavailable, or unreliable. Rather than treating a label or reward as an immutable attribute of a training example, DE-Venus models supervision as state whose content, confidence, and persistence may evolve from data preparation to policy optimization. This abstraction yields three complementary intervention modules. \emph{Active Data Selection} determines which examples should be retained, weakly supervised, or annotated before optimization. \emph{Weak Supervision Construction} turns model-generated evidence into pseudo targets or target-free rewards for examples without trusted answers. \emph{Training-Time Supervision Refinement} reassesses constructed or existing supervision using reward dynamics, representation geometry, and rollout evidence. Their ordering follows the supervision lifecycle, but they are not mandatory consecutive stages: an experiment activates only the interventions required by its supervision setting.

The systems contribution of DE-Venus is a supervision control plane that spans these three intervention points without replacing the distributed RL runtime. DE-Venus owns experiment-level dispatch, method-local supervision transformations, evidence and provenance management, and persistent dataset transitions. verl continues to own data transport, rollout execution, distributed model computation, optimization, validation, and checkpoint services. Minimal invasion therefore does not mean that every method is expressible as a small callback. It means that even when a method specializes controller flow, it preserves the dependency direction and data contracts of the underlying runtime.

\subsection{Design Principles}
\label{subsec:devenus-principles}
DE-Venus follows four principles that turn the supervision lifecycle into a reusable systems abstraction.

\paragraph{Decouple supervision semantics from distributed execution.} Constructing a pseudo target, assigning an intrinsic reward, admitting a reliable trajectory group, or proposing a label revision is a method-level decision. Model placement, rollout generation, distributed forward and backward computation, optimization, and checkpointing are backend responsibilities. DE-Venus introduces control logic only where supervision is interpreted or consumed and delegates the corresponding systems operations to verl.

\paragraph{Preserve native contracts at intervention points.} Minimal invasion is achieved by preserving interfaces rather than forcing every method into one callback. Online methods materialize their decisions as targets, token-level rewards, aligned \texttt{DataProto} selections, or advantages already understood by downstream verl components. Offline selection, promotion, and relabeling materialize \texttt{RLHFDataset}-compatible Parquet datasets, while method-specific evidence and provenance remain separate artifacts. No second worker protocol, data representation, or optimization backend is introduced.

\paragraph{Make supervision lifetimes explicit.} A supervision decision may be local to one optimization step or persistent across training rounds. Step-local decisions replace a target or reward, filter a rollout group, or transform an advantage in memory. Persistent decisions select an example, promote an unlabeled sample, or revise an unreliable target in a versioned dataset. Separating these lifetimes prevents transient rollout evidence from silently mutating persistent supervision and makes every cross-round change auditable.

\paragraph{Control experimental variation.} Hydra configuration and launch workflows expose method selection while retaining a shared execution environment. Whenever algorithmic requirements permit, methods use the same model, dataset loader, rollout engine, GRPO implementation, resource allocation, logging, validation, and checkpoint path. The architecture does not require identical controller flow; it limits infrastructure-level confounders so that comparisons primarily reflect differences in supervision strategy.

\subsection{Architecture Overview}
\label{subsec:devenus-overview}
Figure~\ref{fig:devenus-architecture} organizes DE-Venus into four functional bands and three ownership domains. The \emph{experiment interface} specifies a reproducible run through Hydra configuration, launch workflows, and a training Parquet dataset. The two middle bands form the DE-Venus core. \emph{Supervision orchestration} assembles the runtime context, dispatches online methods, and manages persistent transitions. The \emph{weak-supervision algorithm} band implements the three intervention modules introduced above and factors common voting, verification, reward, uncertainty, and provenance services across them. The bottom \emph{distributed RL runtime} remains owned by verl and provides the dataset protocol, rollout engines, GRPO update, distributed workers, and runtime services.

The horizontal ordering of the three intervention modules describes when supervision may change, not a compulsory pipeline. Active Data Selection is an upstream producer of training data. Weak Supervision Construction operates in the online rollout-to-update path. Training-Time Supervision Refinement may act online by filtering or reweighting trajectories and may also persist accepted promotions or corrections between runs. All three share the same Parquet schema and, whenever they enter online optimization, the same \texttt{DataProto} and worker contracts.

\subsection{Unified Invocation and the verl Boundary}
\label{subsec:devenus-invocation}
DE-Venus adopts verl's configuration and invocation model rather than introducing a parallel control stack \citep{sheng2024hybridflow_verl}. A Hydra overlay inherits the trainer configuration and adds selectors for the supervision setting and method. Model, dataset, rollout, parallelization, optimization, resource, validation, and checkpoint options remain in the verl configuration space. The online methods evaluated in this report use GRPO as a common policy-optimization interface \citep{grpo}.

At runtime, \texttt{TaskRunner} is executed as a Ray remote actor. It resolves the composed configuration and assembles a shared execution context containing worker roles and resource pools, model and data dependencies, reward managers, validation services, datasets, samplers, and collators. This context is injected into \texttt{WeakTrainerInterface}, which dispatches the selected online trainer while preserving the standard \texttt{init\_workers()} and \texttt{fit()} lifecycle. Cluster construction and model placement therefore remain independent of method selection.

The dispatch boundary intentionally covers online computation only. Active Data Selection executes before optimization, while persistent promotion and relabeling execute between training rounds. These offline components emit a common Parquet dataset and provenance artifacts, and a subsequent invocation consumes the result through the ordinary \texttt{RLHFDataset} path. Input-shape specializations follow the same placement rule. For example, Co-rewarding aligns original and rewritten questions during dataset construction, the highest layer that understands their pairing, and then reuses the standard rollout and optimization interfaces.

Minimal invasion is consequently a property of this boundary rather than a claim about method code size. A trainer may require paired rollout passes, periodic probes, reliability filtering, or a specialized ordering of reward and advantage computation. Nevertheless, method code invokes existing worker APIs and produces objects expected by the next verl stage; it neither reconstructs the cluster nor forks the distributed execution backend.

\subsection{Three Supervision Modules}
\label{subsec:devenus-modules}
The three modules share one execution substrate but differ in decision time, evidence source, and boundary object. Table~\ref{tab:devenus-modules} summarizes their implementation forms. This architectural grouping follows where a method intervenes in the supervision lifecycle; individual trainers may still be organized by semi-supervised, unsupervised, or noisy-label assumptions.

\begin{table*}[t]
\centering
\small
\setlength{\tabcolsep}{6pt}
\renewcommand{\arraystretch}{1.12}
\begin{tabularx}{\textwidth}{>{\raggedright\arraybackslash}p{0.29\textwidth}>{\raggedright\arraybackslash}p{0.39\textwidth}>{\raggedright\arraybackslash}X}
\toprule
\textbf{Intervention module} & \textbf{Execution form} & \textbf{Boundary object} \\
\midrule
Active Data Selection & Offline generate--score--route pipeline & Training Parquet \\
Weak Supervision Construction & Method-local online trainer & Target or reward fields \\
Training-Time Supervision Refinement & Online filtering with optional cross-round transition & Aligned batch or revised Parquet \\
\bottomrule
\end{tabularx}
\caption{Execution forms and boundary objects of the three DE-Venus intervention modules.}
\label{tab:devenus-modules}
\end{table*}

\subsubsection{Active Data Selection}
\label{subsubsec:devenus-selection}
Active Data Selection is implemented as an offline generate--score--route pipeline and is not dispatched through \texttt{WeakTrainerInterface}. A response generator first performs multi-sample inference over the candidate pool. Scoring adapters then aggregate response behavior and model-internal statistics into question-level signals such as empirical accuracy, consistency, entropy, self-certainty, attention pivots, and hidden-state dynamics. A selector consumes these normalized records together with annotation and training budgets and determines the destination of each candidate.

The decision procedure supports both verified and initially unlabeled pools. When answers are available, empirical pass rate restricts the admissible difficulty region and an uncertainty signal ranks the remaining examples. When answers are unavailable, the selector ranks the pool by model uncertainty, samples a small set of probes across that ranking, and uses their annotations to calibrate data-dependent routing thresholds. The resulting routes distinguish examples that can be excluded, retained for weakly supervised training, or prioritized for annotation. This offline placement keeps candidate generation and calibration artifacts outside the distributed training protocol.

The final materializer writes the selected examples directly into the common Parquet schema consumed by \texttt{RLHFDataset}. Each retained record carries its prompt, current target, label state, stable sample key, source, and task metadata. Consequently, the selected dataset can be combined with any compatible online intervention by changing the input Parquet alone.

\subsubsection{Weak Supervision Construction}
\label{subsubsec:devenus-construction}
Weak Supervision Construction is implemented by method-local online trainers selected via \texttt{WeakTrainer}\allowbreak\texttt{Interface}. Each trainer receives the execution context assembled by \texttt{TaskRunner} and specializes controller-side logic around the inherited \texttt{RayPPOTrainer} lifecycle. The principal intervention occurs after grouped rollouts are available and before the corresponding reward or advantage consumer, so supervision can be derived from multiple responses without altering generation workers.

The module supports two output forms. Pseudo-target methods infer a temporary answer from response consensus or agreement across paired views and write it to \path{reward_model.ground_truth} before verifier scoring. TTRL uses within-question consensus, whereas Co-rewarding coordinates original and rewritten questions to introduce cross-view evidence. Target-free methods instead convert confidence or uncertainty statistics into sequence-level supervision and write the resulting signal through the reward interface. Intuitor and EMRL instantiate this path with self-certainty and entropy, respectively. The unsupervised trainer uses the same reward interface for confidence, agreement, and self-verification signals. In a semi-supervised batch, verified targets remain active for labeled examples while constructed supervision is applied only where trusted answers are unavailable.

Both forms terminate at a native verl contract: either the verifier receives a substituted target or GRPO receives a response-shaped reward tensor. Once that contract has been materialized, advantage estimation, actor update, metric reduction, validation, and checkpointing proceed through the shared runtime.

\subsubsection{Training-Time Supervision Refinement}
\label{subsubsec:devenus-refinement}
Training-Time Supervision Refinement reassesses supervision after rollout evidence becomes available. Its online path admits or reweights trajectories before actor optimization. TraPO tracks reward and pass-rate dynamics over training and uses their temporal behavior to retain reliable supervision; GeoMin evaluates rollout groups against representation distributions estimated from verified examples and uses the resulting geometric evidence for filtering or advantage transformation. These operations apply one aligned selection or tensor transformation to the complete \texttt{DataProto}, preserving the correspondence between trajectories and metadata.

The module also supports decisions whose effects must persist beyond the current run. Periodic probe passes collect majority answers, pass rates, and trajectory histories without performing actor updates. TraPO can use this evidence to promote reliable unlabeled examples, while OLR compares a stable model-generated candidate with the existing annotation and proposes a correction only when its trajectory evidence satisfies the configured reliability rule. Probe records are keyed by stable sample identity rather than by transient rollout groups, allowing observations from different steps and rounds to be joined consistently.

Accepted promotions and corrections are materialized outside the trainer and Ray worker graph. An offline transition builder consumes the current Parquet and persisted evidence, applies the decision rule, and writes both the next-round dataset and a manifest of accepted changes. Distributed workers therefore provide evidence but never mutate persistent labels, and the revised dataset re-enters training through the same invocation and dataset contracts as its predecessor.

\subsection{Unified Data Model}
\label{subsec:devenus-data-model}
DE-Venus uses a two-level data model to connect cross-round dataset evolution with within-step RL optimization. The persistent level represents the current supervision state of each sample, whereas the runtime level represents trajectories generated from that sample in one optimization step. Let the training dataset at round $k$ be
\begin{equation}
\mathcal{D}^{(k)}=\left\{d_s^{(k)}=\left(x_s,y_s^{(k)},z_s^{(k)},m_s\right)\;\middle|\;s\in\mathcal{I}^{(k)}\right\},
\end{equation}
where $s$ is the persistent sample key stored in \path{extra_info.index}, and $\mathcal{I}^{(k)}$ is the set of samples present in round $k$. The remaining components denote the prompt $x_s$, supervision target $y_s^{(k)}$ in \path{reward_model.ground_truth}, label state $z_s^{(k)}$ in \path{extra_info.labeled}, and task metadata $m_s$, including \texttt{data\_source} and \texttt{ability}. Selection may change membership in $\mathcal{I}^{(k)}$, while promotion or correction may change $y_s^{(k)}$ and $z_s^{(k)}$, without changing the identity of a retained sample.

At ingestion, \texttt{RLHFDataset} exposes each persistent record to the dataloader and copies $s$ to the batch-level \texttt{index}. The collated batch is represented as a \texttt{DataProto}: its tensor partition carries model inputs and optimization quantities, while its non-tensor partition carries targets, label states, task metadata, and persistent keys. Controller-side transformations operate on this composite object so semantic metadata remains aligned with the corresponding trajectories. Offline modules read or materialize $\mathcal{D}^{(k)}$, whereas online trainers operate on its in-memory \texttt{DataProto} representation.

For optimization step $t$, let $\mathcal{I}_t\subseteq\mathcal{I}^{(k)}$ denote the sampled records. The trainer assigns a fresh group identifier $u_s^{(t)}$ to each $s\in\mathcal{I}_t$ and repeats the prompt $n$ times before generation. The resulting rollout group is
\begin{equation}
\mathcal{G}_{s}^{(t)}=\left\{\tau_{s,j}^{(t)}\right\}_{j=1}^{n},\qquad\operatorname{uid}\!\left(\tau_{s,j}^{(t)}\right)=u_s^{(t)}.
\end{equation}
The shared \texttt{uid} lets groupwise operations, including majority voting and GRPO normalization, recover sibling trajectories after batch balancing or reordering. Unlike the persistent key $s$, $u_s^{(t)}$ is local to the current rollout batch and is regenerated at later steps. Probe histories, transition manifests, and successive Parquet datasets therefore join records by \path{extra_info.index}, never by \texttt{uid}.

\subsection{Contract-Preserving Dataflows}
\label{subsec:devenus-dataflows}
DE-Venus represents an online method as a composition of optional supervision transformations around verl's rollout-to-update path. For optimization step $t$, the dataflow is
\begin{equation}
\begin{aligned}
\mathcal{B}_{t} &\xrightarrow{\text{rollout}} \mathcal{T}_{t}
\xrightarrow{\mathcal{S}_{m}^{y}} \mathcal{T}_{t}^{y}, \\
\mathcal{T}_{t}^{y} &\xrightarrow{\mathcal{S}_{m}^{b}} \widetilde{\mathcal{T}}_{t}
\xrightarrow{\text{reward}} R_t
\xrightarrow{\mathcal{S}_{m}^{r}} \widetilde{R}_t, \\
\widetilde{R}_t &\xrightarrow{\text{GRPO}} A_t
\xrightarrow{\mathcal{S}_{m}^{a}} \widetilde{A}_t
\xrightarrow{\text{update}} \theta_{t+1}.
\end{aligned}
\label{eq:devenus-online-dataflow}
\end{equation}
Here $\mathcal{S}_{m}^{y}$, $\mathcal{S}_{m}^{b}$, $\mathcal{S}_{m}^{r}$, and $\mathcal{S}_{m}^{a}$ denote target construction, batch admission, reward construction, and advantage transformation for method $m$, respectively. An unused transformation is the identity. Each active transformation is placed after its required evidence becomes available and immediately before the first downstream stage that consumes its output. Table~\ref{tab:devenus-contracts} summarizes the corresponding runtime contracts.

\begin{table*}[t]
\centering
\small
\setlength{\tabcolsep}{6pt}
\renewcommand{\arraystretch}{1.12}
\begin{tabularx}{\textwidth}{>{\raggedright\arraybackslash}p{0.18\textwidth}>{\raggedright\arraybackslash}p{0.34\textwidth}>{\raggedright\arraybackslash}X}
\toprule
\textbf{Surface} & \textbf{Native contract} & \textbf{Method operation} \\
\midrule
Target & \path{reward_model.ground_truth} & Pseudo-target substitution \\
Reward & \texttt{token\_level\_scores} & Intrinsic or proxy signal \\
Batch & Aligned \texttt{DataProto} & Trajectory-group admission \\
Advantage & \texttt{advantages} & Reliability-aware reweighting \\
\bottomrule
\end{tabularx}
\caption{Online intervention surfaces and native verl contracts preserved by DE-Venus.}
\label{tab:devenus-contracts}
\end{table*}

Contract preservation is enforced by two invariants. First, when a method produces a sequence-level score, DE-Venus places it at the final valid response token, yielding the response-shaped reward tensor expected by verl. Second, admission and filtering use aligned \texttt{DataProto} selections, so tensor and non-tensor partitions retain the same row mapping. After the final active transformation, the remaining verl stages execute unchanged.

Decisions that must survive the current optimization step follow a separate outer dataflow:
\begin{equation}
\begin{aligned}
\mathcal{D}^{(k)} &\xrightarrow{\text{collect}} \mathcal{E}^{(k)}
\xrightarrow{\Pi_m} \Delta_m^{(k)}, \\
\Delta_m^{(k)} &\xrightarrow{\text{materialize}} \mathcal{D}^{(k+1)},
\end{aligned}
\label{eq:devenus-persistent-dataflow}
\end{equation}
where $\mathcal{E}^{(k)}$ contains generated responses, uncertainty scores, or index-keyed probe evidence; $\Pi_m$ is the method-specific decision rule; and $\Delta_m^{(k)}$ is the accepted set of selections, promotions, or target revisions. Active Data Selection applies this pattern to a candidate pool to construct $\mathcal{D}^{(0)}$; TraPO applies it to promote reliable unlabeled samples; and OLR applies it to revise unreliable targets. Probe-only rollouts terminate in $\mathcal{E}^{(k)}$ and do not enter actor optimization. Each dataset transition is therefore reproducible from its source dataset, persisted evidence, decision configuration, and manifest.

\subsection{Diagnostics, State, and Extensibility}
\label{subsec:devenus-extensibility}
\paragraph{Supervision observability.} Weakly supervised optimization must expose the quality of its supervision decisions, not only final task reward. DE-Venus augments verl's optimization and systems metrics with pseudo-target agreement, pass-rate and reward dynamics, admitted-group counts, confidence estimates, and reward and advantage statistics. These diagnostics enter the same step-indexed tracking and validation path as native verl metrics. Persistent decisions additionally retain response or probe records and transition manifests, providing a sample-level audit trail for selection, promotion, and correction.

\paragraph{State ownership.} DE-Venus assigns state according to lifetime. Step-local supervision state remains in \texttt{DataProto} and is discarded after the update. Run-resumable decision state extends the inherited checkpoint only when later behavior depends on it. TraPO, for example, persists majority answers, pass-rate histories, probe counters, and associated decision buffers alongside the normal training state. Cross-round state is externalized as Parquet datasets, evidence records, and manifests keyed by persistent sample identity; it is never reconstructed from an ephemeral \texttt{uid} or worker-local memory.

\paragraph{Extension protocol.} An online method reuses the execution context assembled by \texttt{TaskRunner} and implements one or more intervention surfaces from Table~\ref{tab:devenus-contracts} by specializing \texttt{RayPPOTrainer} or an existing DE-Venus trainer. Registration with \texttt{WeakTrainerInterface} exposes it through the common Hydra entry point. An offline method instead implements the evidence--decision--materialization dataflow in Equation~\ref{eq:devenus-persistent-dataflow} and emits an \texttt{RLHFDataset}-compatible Parquet dataset with the required evidence or provenance. Neither extension requires a new worker role, Ray protocol, parallelization strategy, or optimizer lifecycle. This protocol operationalizes minimal invasion while allowing method-specific control over supervision.

%% file: sections/method.tex
\section{Data-Efficient RLVR: From Data Curation to Reliable Supervision}\label{sec:methods}
Reinforcement learning with verifiable rewards (RLVR) typically relies on training instances whose responses can be evaluated against trustworthy ground-truth answers. This dependence creates a fundamental data bottleneck: obtaining verified labels is costly, uniformly using all available examples may waste annotation and rollout budgets, and large collections of unlabeled data cannot directly provide verifiable rewards. Moreover, both automatically constructed supervision and existing annotations may be unreliable, causing noisy or even harmful policy updates. Data-efficient RLVR therefore aims to extract effective learning signals from limited and imperfect supervision while controlling the quality of the resulting rewards.
We organize the methods supported by our framework into three stages. First, \emph{active data selection} estimates the utility of candidate examples and allocates limited annotation and training budgets to informative data. Second, \emph{weak supervision construction} enables examples without trusted answers to participate in RLVR by constructing either pseudo targets or intrinsic rewards. Third, \emph{training-time supervision refinement} filters, reweights, or corrects imperfect supervision before it affects policy optimization. This organization provides a unified view of semi-supervised, label-free, and noisy-label RLVR: they differ primarily in the availability and reliability of the supervision entering these three stages. 
After active data selection, let
\(\mathcal D_{\ell}\) and \(\mathcal D_u\) denote the selected
labeled and unlabeled training sets, respectively. Given a
question \(x_i\) and its \(k\)-th rollout
\(y_{i,k}\sim\pi_\theta(\cdot\mid x_i)\), we formulate the
unified training interface as
\begin{equation}
r_{i,k}=
\begin{cases}
R(y_{i,k},a_i^\star), & (x_i,a_i^\star)\in\mathcal D_{\ell},\\
R_u(y_{i,k};x_i), & x_i\in\mathcal D_u,
\end{cases}
\qquad
\widetilde A_{i,k}=w_i\widehat A_{i,k},\quad w_i\in[0,1].
\label{eq:unified_weak_supervision}
\end{equation}
Here,
\(R(y,a)=\mathbb I[\operatorname{Ans}(y)=a]\) denotes the
standard verified reward, while \(R_u\) denotes the weak reward
constructed for an unlabeled question. Specifically, \(R_u\)
can be instantiated either as a pseudo-target reward,
\(\mathbb I[\operatorname{Ans}(y_{i,k})=\tilde a_i]\), or as a
target-free intrinsic reward
\(r_{\mathrm{int}}(x_i,y_{i,k};\theta)\).
The standard group-normalized advantage
\(\widehat A_{i,k}\) is further modulated by the reliability
weight \(w_i\). Accordingly, training-time refinement can filter
a question by setting \(w_i=0\), softly reweight its supervision
with \(0<w_i<1\), or correct the pseudo target \(\tilde a_i\)
used by \(R_u\).

\begin{figure*}[t]
    \centering
    \includegraphics[width=\textwidth]%
    {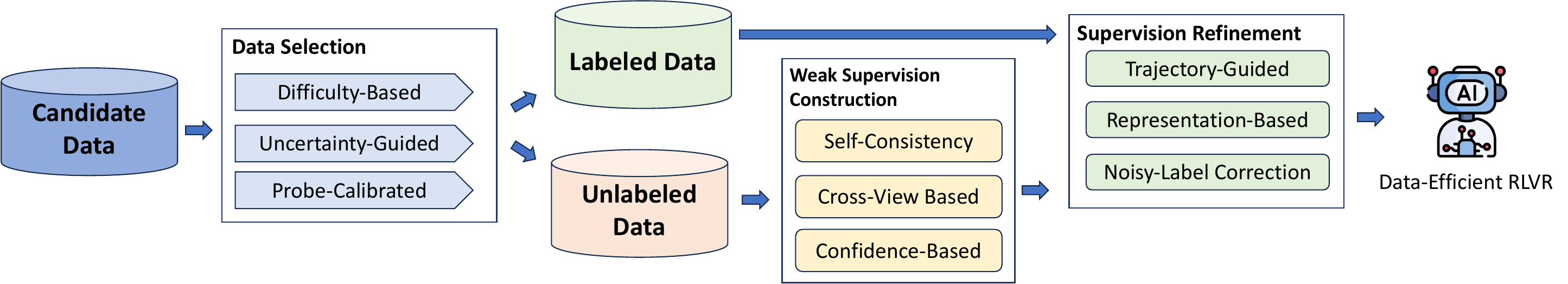}
    \caption{
        Overview of the weakly supervised RLVR pipeline.
        Candidate data are first selected before training, followed by weak
        supervision construction for selected unlabeled samples and
        supervision refinement during policy optimization.
        Potentially noisy labels are handled through a separate
        training-time correction branch.
    }
    \label{fig:weak-supervision-pipeline}
\end{figure*}

\subsection{Active Data Selection}
\label{sec:pre_training_selection}

Active data selection determines how annotation and optimization resources are allocated before RLVR training begins. Because candidate questions differ in both learning utility and their need for external supervision, selection is not limited to retaining a smaller training subset; it may also determine which questions should be annotated, trained with weak supervision, or excluded from training. Depending on the supervision available during selection, the methods supported by our framework fall into three categories: difficulty-based filtering using verified outcomes, uncertainty-guided selection using model-derived signals, and probe-calibrated data triage using a small amount of targeted annotation. Together, these categories progress from accurate but label-dependent selection to annotation-efficient selection from initially unlabeled data.

\subsubsection{Difficulty-Based Data Filtering}
\label{sec:difficulty_selection}

Difficulty-based filtering provides the most direct way to estimate the learning utility of a question when verified answers are available. Existing RLVR systems commonly sample multiple responses from the current policy and use their empirical pass rate as a model-dependent measure of question difficulty. Questions that are consistently solved are regarded as already mastered, whereas questions for which the policy rarely produces a correct response may provide insufficient reward variation for effective optimization; consequently, difficulty-aware methods typically retain questions within an intermediate pass-rate range \citep{yu2025dapo,bae2025online,zeng2025cures}. Our framework implements this paradigm through configurable accuracy intervals, after which the retained questions can be ranked by an auxiliary uncertainty criterion when they exceed the available training budget. This approach provides an interpretable mechanism for controlling both sample difficulty and training-set size, but it assumes that verified answers are already available for the entire candidate pool. When such full annotation is unavailable, data utility must instead be inferred from signals produced by the model itself.

\subsubsection{Uncertainty-Guided Data Selection}
\label{sec:uncertainty_selection}

Uncertainty-guided selection replaces verified correctness with question-level signals derived from model outputs or internal representations. Response-based methods use agreement among multiple sampled answers, treating lower consistency as higher uncertainty, while probability-based methods quantify uncertainty through statistics such as token-level entropy or self-certainty. Representation-based methods further examine the internal dynamics of reasoning: CoE characterizes layer-wise changes in hidden-state magnitude and direction, whereas CoT-Kinetics describes the semantic movement and curvature of reasoning representations across layers \citep{zuo2025ttrl,huang2023uncertainty,tang2025towards,kang2026selfcertainty,wang2024latent,bi2025cot}. Our framework exposes consistency, entropy, self-certainty, CoE, and CoT-Kinetics as interchangeable selection strategies, each of which can rank the candidate pool and select questions under a predefined threshold or budget. Although some of these signals are also adopted by weakly supervised RLVR algorithms, their role here is restricted to deciding which questions enter the training pool rather than constructing rewards for individual responses. These methods remove the need to annotate every candidate question, but their raw scores may not be comparably calibrated across models and datasets, motivating the use of limited annotations to determine data-dependent selection boundaries.

\subsubsection{Probe-Calibrated Data Triage}
\label{sec:probe_calibrated_selection}

Probe-calibrated data triage addresses this calibration problem by combining model-derived uncertainty with a small, strategically distributed set of annotations. It first ranks the initially unlabeled question pool using an uncertainty estimator, samples probe questions across the resulting spectrum, and annotates only these probes to estimate how the selection score relates to the current policy's empirical correctness. The estimated relationship is then used to determine thresholds for different data routes. PivotTrace is a representative method in this category: it detects metacognitive pivots from long-range attention patterns in generated reasoning trajectories, uses the number of detected pivots as an uncertainty proxy, and calibrates the pivot-count ranking through sliding-window statistics over a small probing set \citep{zhu2026pivottrace}. Based on the calibrated thresholds, low-uncertainty questions that have likely been mastered are discarded, questions with intermediate uncertainty are retained for weakly supervised training, and highly uncertain questions are prioritized for external annotation. Our framework generalizes this procedure by allowing consistency, entropy, self-certainty, CoE, and CoT-Kinetics to replace pivot count as the underlying ranking criterion. The resulting data partition completes the selection stage: verified answers are available for the annotated subset, while the retained unlabeled subset is passed to the subsequent supervision-construction stage.

\subsection{Weak Supervision Construction}
\label{sec:weak_supervision_construction}

Following active data selection, the remaining challenge is to make unlabeled questions usable for RLVR. Unlike labeled examples, these questions lack verified answers against which generated responses can be evaluated. Weak supervision construction addresses this missing-verifier problem by converting model-generated information into response-level learning signals. According to the source and form of the constructed supervision, the methods supported by our framework fall into three categories: self-consistency methods that infer pseudo targets from repeated responses, cross-view methods that construct supervision from complementary sources, and confidence-based methods that directly assign rewards without producing an explicit target. These categories progressively move from reconstructing a temporary verifier to bypassing answer-based verification altogether.

\subsubsection{Self-Consistency-Based Pseudo-Target Construction}
\label{sec:self_consistency_supervision}

Self-consistency-based methods approximate a missing verifier through the collective prediction of the current policy. TTRL samples multiple responses for each unlabeled question, groups their extracted answers, and treats the majority answer as a pseudo target \citep{zuo2025ttrl}. Individual responses can then be evaluated by whether their answers agree with this target, allowing the resulting binary signals to replace ground-truth rewards during policy optimization. Our framework supports this construction through both TTRL and ensemble-style voting over multiple rollouts, while preserving verified rewards for the labeled portion of a semi-supervised dataset. This approach is simple and compatible with the standard RLVR pipeline because it ultimately recovers an answer-based verifier. However, responses sampled from the same policy may exhibit correlated errors and converge on a shared but incorrect answer, causing majority voting to reinforce an internally consistent mistake. This weakness motivates the use of complementary views when constructing pseudo supervision.

\subsubsection{Cross-View Agreement-Based Supervision}
\label{sec:cross_view_supervision}

Cross-view methods strengthen pseudo supervision by requiring agreement across complementary views rather than relying exclusively on repeated responses to the same question. Co-Rewarding instantiates this idea by rewriting each question into a semantically equivalent variant, generating responses for both the original and rewritten questions, and constructing supervision from cross-view agreement \citep{zhang2025corewarding}. Because the two views express the same underlying reasoning problem through different surface forms, their agreement provides an additional constraint on whether a candidate answer should be reinforced. Our framework implements this data-side construction by coordinating rollouts and cross-voting between the original and rewritten questions, thereby extending single-view majority voting without requiring external annotations. Nevertheless, both self-consistency and cross-view agreement ultimately commit to a discrete pseudo target, whose correctness still depends on the quality and diversity of the model-generated candidates. A different family of methods avoids this commitment by deriving rewards directly from the policy's predictive distributions.

\subsubsection{Confidence-Based Target-Free Rewards}
\label{sec:target_free_rewards}

Confidence-based methods bypass pseudo-target construction and directly use the model's internal uncertainty as a reward signal. Intuitor adopts self-certainty as its sole supervision, rewarding responses for which the policy assigns a more concentrated predictive distribution \citep{zhao2025learning}. EM-RL follows a closely related principle by using negative token- or sequence-level entropy as the reward, thereby encouraging the policy to reinforce responses generated with greater confidence \citep{agarwal2025unreasonable}. Our framework implements this family through configurable self-certainty, token-level entropy, trajectory-level entropy, and sequence-probability rewards, which can be computed for each response and directly incorporated into group-relative advantage estimation. Although similar statistics also appear in active data selection, their roles are different: selection aggregates them at the question level to determine which data enter training, whereas target-free reward construction applies them at the response level to guide policy updates. By eliminating both verified answers and discrete pseudo targets, these methods provide a broadly applicable source of internal supervision; however, model confidence does not necessarily imply correctness, and an overconfident policy may reinforce its own errors. This remaining reliability problem motivates the training-time supervision refinement methods introduced next.

\subsection{Training-Time Supervision Refinement}
\label{sec:training_time_refinement}

Weak supervision should not be treated as a fixed training signal because its reliability may change as the policy evolves. Supervision entering RLVR can be imperfect for two complementary reasons: unlabeled questions rely on pseudo targets whose correctness is initially uncertain, while labeled questions may contain annotations that are themselves incorrect. Training-time supervision refinement uses evidence collected during optimization or between successive training rounds to reassess these signals and determine whether they should be retained, promoted, reweighted, or replaced. We organize the supported methods according to the evidence used for this decision: learning trajectories track how candidate supervision evolves over time, representation geometry measures its compatibility with reliable rollouts, and online label correction resolves conflicts between model-generated candidates and existing annotations.

\subsubsection{Trajectory-Guided Pseudo-Label Refinement}
\label{sec:trajectory_refinement}

Trajectory-guided refinement evaluates pseudo labels through their behavior over the course of policy optimization. TraPO uses a small labeled subset to establish reliable learning trajectories and retains unlabeled examples whose pseudo-label trajectories exhibit similar dynamics \citep{yang2025trapo}. In particular, it tracks the pass rate of majority-voted answers across training steps and compares their temporal patterns with those observed on verified examples, allowing pseudo labels that behave inconsistently with labeled supervision to be excluded from policy updates. Our framework supports this mechanism through the original TraPO filtering procedure and further extends it with TraPO v2, which performs labeled-only warmup, periodically probes the candidate pool, and promotes an unlabeled example only when its current majority answer achieves a sufficiently high pass rate, a positive growth trend, and stable historical consistency. TraPO v2 therefore turns reliability estimation into an iterative process in which the trusted subset expands as the policy develops. However, trajectory-based evidence requires repeated observations before a decision can be made, motivating complementary approaches that assess pseudo-label reliability through the policy's internal representation structure.

\subsubsection{Representation-Based Reliability Filtering}
\label{sec:representation_refinement}

Representation-based filtering determines pseudo-label reliability by comparing unlabeled rollouts with the geometric structure learned from verified examples. GeoMin models the layer-wise hidden representations of correct and incorrect labeled rollouts using von Mises--Fisher distributions and identifies the layers that most clearly separate the two groups \citep{zhu2026geomin}. For an unlabeled question, it first obtains a majority pseudo target and then evaluates whether the corresponding majority and minority rollouts align with the correct and incorrect representation distributions, respectively. The resulting distributional affinity is used as a confidence signal, while a two-component Gaussian mixture model adaptively separates reliable pseudo labels from unreliable ones. Our framework uses this signal to filter pseudo-labeled examples and to reweight selected ambiguous updates during training. Although GeoMin performs sample filtering, its input consists of pseudo-labeled rollouts generated during optimization rather than raw questions before training, placing it in supervision refinement rather than active data selection. Both GeoMin and TraPO thus address weak supervision caused by missing labels; yet supervision may also be weak because an apparently verified label is incorrect. In this latter setting, filtering constructed pseudo labels is insufficient, and refinement must instead decide when an existing target should be replaced.

\subsubsection{Online Noisy-Label Correction}
\label{sec:noisy_label_refinement}

Online noisy-label correction addresses this complementary setting by refining labels that are available but potentially corrupted. In RLVR, the effect of a label depends on whether the current policy can generate responses that match it, making rollout dynamics a natural source of evidence for identifying unreliable annotations. Online Label Refinement (OLR) therefore preserves the provided labels during an initial learning phase and subsequently tracks the majority answer generated for each question across policy updates \citep{yang2026can}. A candidate answer is considered for correction only when its rollout pass rate exhibits a positive trend and its identity remains consistent across recent training history, reducing the risk of replacing a label with a transient model prediction. The NL-OLR v2 implementation in our framework further compares the candidate trajectory with that of the current label and performs relabeling only when the candidate satisfies absolute reliability thresholds and improves over the existing target by a configurable margin. Whereas TraPO v2 promotes previously unlabeled examples into the trusted subset, NL-OLR v2 revises supervision for examples that were already labeled. Together with representation-based filtering, these methods allow the framework to refine weak supervision arising from both missing and noisy annotations as training progresses.

%% file: sections/derl_experiments.tex

\section{Experiments}
\label{sec:derl_experiments}

This section evaluates DE-Venus along three complementary dimensions of data-efficient RLVR: learning from incomplete supervision, robust training with corrupted supervision, and allocation of annotation and training budgets through data selection. We first conduct controlled studies on public reasoning benchmarks and then validate the same framework capabilities in three business scenarios. All compared methods are instantiated through the common DE-Venus training and evaluation pipeline, with variation confined to the supervision or selection strategy under study. The experiments are intended to characterize framework coverage and practical behavior rather than to claim the individual methods as new contributions of DE-Venus.

\paragraph{Optimization protocol.}
Unless otherwise stated, the reported RLVR experiments instantiate DE-Venus with GRPO~\citep{grpo} as the default policy-optimization backbone. Holding the optimization rule fixed allows observed differences to be attributed primarily to data selection, weak-supervision construction, and label refinement rather than to the choice of RL algorithm. This protocol does not define the capability boundary of the framework: because DE-Venus preserves verl's native reward, batch, and advantage interfaces, the same supervision modules can be configured with other commonly used RLVR formulations, including RLOO~\citep{ahmadian2024back} and REINFORCE++~\citep{hu2025reinforcepp}, among others. A systematic comparison across RLVR algorithms is beyond the scope of this report.

\paragraph{Shared evaluation protocol.}
The public studies evaluate in-domain (ID) mathematical reasoning on AIME 2024, AIME 2025, AMC~\citep{li2024numinamath}, MATH-500~\citep{hendrycks2021measuring}, Minerva~\citep{lewkowycz2022solving}, and OlympiadBench~\citep{he2024olympiadbench}, and out-of-domain (OOD) general reasoning on ARC-c~\citep{clark2018think}, GPQA-Diamond~\citep{rein2024gpqa}, and MMLU-Pro~\citep{wang2024mmlu}. We report $\operatorname{avg}@32$ for AIME 2024/2025 and AMC, $\operatorname{pass}@1$ for MMLU-Pro, and $\operatorname{avg}@4$ for the remaining benchmarks. Decoding uses temperature~0.6 and top-$p=1.0$. Because the three studies use different backbones and training corpora, results should be compared within each subsection rather than across them.

\subsection{Weakly Supervised Learning}
\label{sec:exp_weak_supervision}

The weak-supervision evaluation compares unsupervised and semi-supervised reinforcement learning with verifiable rewards (RLVR). It measures how different reward and sample-utilization strategies perform when labels are unavailable or when only $10\%$ of the training examples are labeled.

\subsubsection{Experimental Setup}
\label{sec:weak_supervision_setup}

\paragraph{Training configuration.}
Training uses the subset of DeepMath-103K~\citep{he2025deepmath} with difficulty scores of at least~6 and Qwen3-8B-Base~\citep{yang2025qwen3} as the backbone. The reported runs use $8\times$A100 GPUs, a total batch size of 128, a micro-batch size of 32, a learning rate of $10^{-6}$, and $G=8$ rollouts per prompt.

\paragraph{Compared methods.}
The comparison includes seven weakly supervised RLVR methods: \textbf{TTRL}~\citep{zuo2025ttrl}, which rewards rollouts that produce the majority answer; \textbf{Tok-entropy} and \textbf{Seq-entropy}~\citep{agarwal2025unreasonable}, which rank rollouts using token-level entropy and sequence probability, respectively; \textbf{Self-certainty}~\citep{zhao2025learning}, which uses the KL divergence between the token distribution and a uniform distribution; \textbf{Co-rewarding}~\citep{zhang2025co}, which uses pseudo-labels generated by a slowly updated reference teacher; \textbf{TraPO}~\citep{yang2025trapo}, which selects unlabeled examples whose pass-rate trajectories are close to those of labeled examples; and \textbf{GeoMin}~\citep{zhu2026geomin}, which combines boundary-focused representation separation with distribution-matched selection of unlabeled examples. TTRL, Tok-entropy, Seq-entropy, Self-certainty, and Co-rewarding are evaluated both without labeled data and with $10\%$ labeled data. TraPO and GeoMin are reported in the latter semi-supervised setting. For all semi-supervised methods, labeled examples receive correctness-based rewards, whereas unlabeled examples use the method-specific self-guided reward.

\begin{figure*}[t]
\centering
\includegraphics[width=0.99\textwidth]{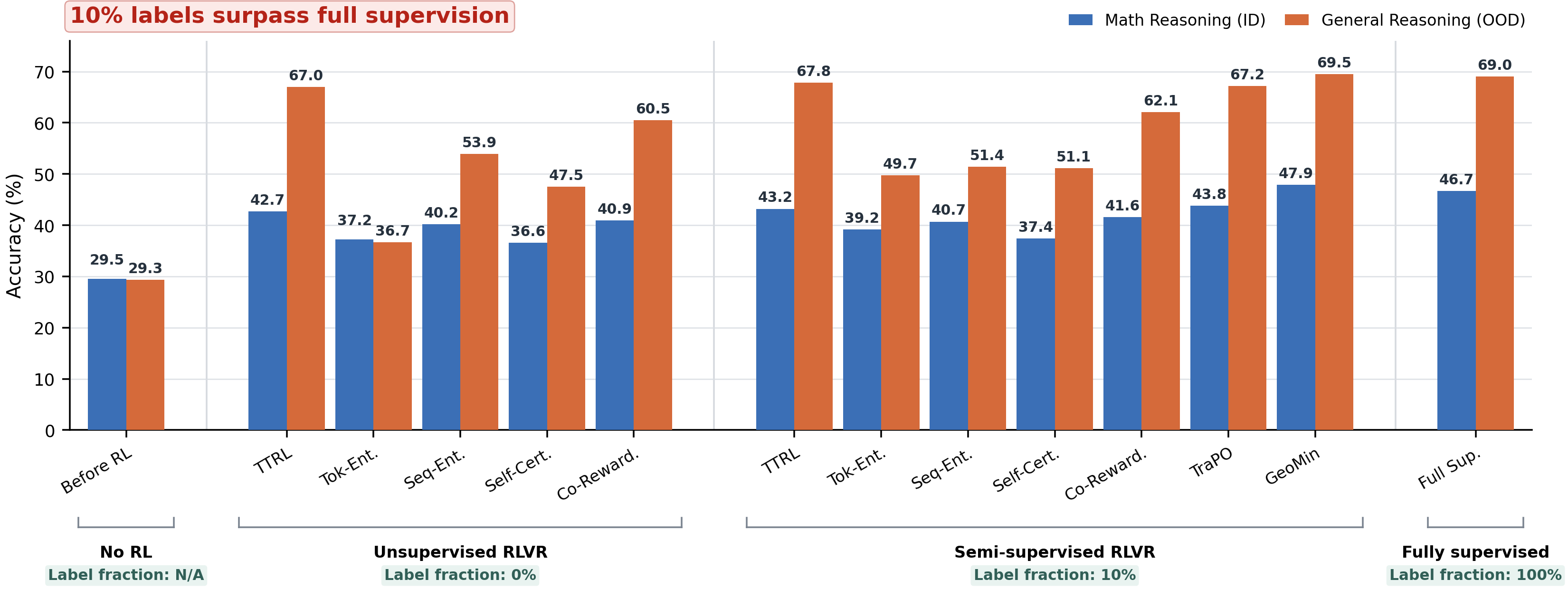}
\caption{\small ID and OOD average accuracy of Qwen3-8B-Base across supervision regimes. The best method trained with 10\% labeled data surpasses the fully supervised baseline on both evaluation groups.}
\label{fig:wsl_results}
\end{figure*}

\subsubsection{Results and Discussion}
\label{sec:weak_supervision_results}

Figure~\ref{fig:wsl_results} shows that useful learning signals can be constructed even without labeled examples: TTRL obtains the strongest unsupervised result, reaching $42.7\%$ ID and $67.0\%$ OOD average accuracy. The wider spread on OOD benchmarks indicates that the choice of self-guided supervision is especially consequential for cross-domain generalization.

The main result appears in the $10\%$-label regime. GeoMin reaches $47.9\%$ ID and $69.5\%$ OOD accuracy, compared with $46.7\%$ and $69.0\%$ under full supervision. It therefore exceeds the fully supervised reference by 1.2 ID points and 0.5 OOD points while using only one tenth of the labels. TraPO also remains competitive at $43.8\%$ ID and $67.2\%$ OOD. Together, these results show that DE-Venus supports both label-free reward construction and selective use of unlabeled examples, and that reliability-aware supervision can recover or surpass full-supervision performance under a sharply reduced annotation budget.

\subsection{Learning with Noisy Labels}
\label{sec:exp_noisy_labels}

The noisy-label evaluation measures the behavior of standard and noise-robustness-enhanced RLVR configurations when a fraction of the supervision is corrupted. The framework varies the noise ratio and supports both inactive and active noisy-label regimes.

\subsubsection{Experimental Setup}
\label{sec:noisy_label_setup}

Experiments use Qwen3-4B-Base and compare standard GRPO with GRPO augmented by online label refinement (OLR). We vary the noise ratio over $\rho\in\{0.1,0.3,0.5,0.7,0.9\}$ under both inactive and active noise. Inactive noisy labels are synthetic targets that the current policy is unlikely to generate, whereas active noisy labels are constructed from on-policy incorrect answers and can therefore be reinforced by sampled rollouts. Every OLR result is compared with standard GRPO at the same noise ratio and under the same noise regime.
\begin{figure*}[t]
\centering
\includegraphics[width=0.99\textwidth]{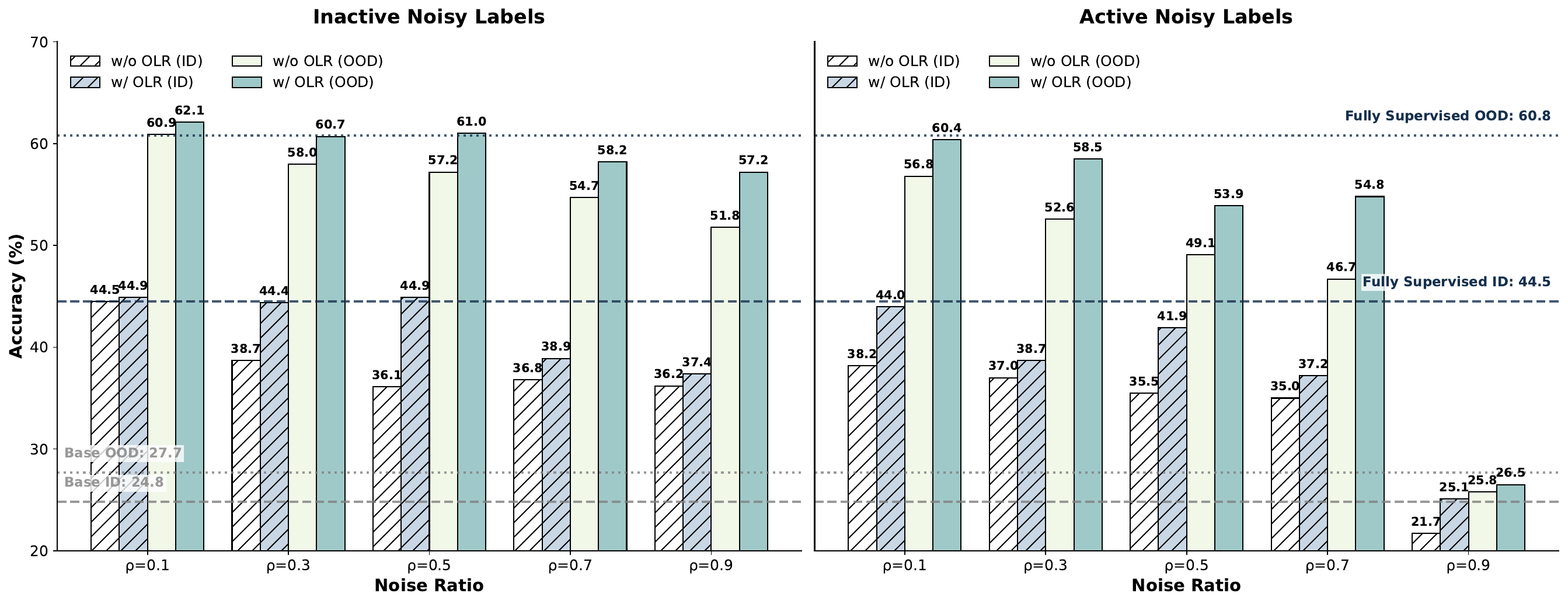}
\caption{ID and OOD average accuracy of Qwen3-4B-Base across label-noise ratios. OLR is compared with standard GRPO under matched inactive- and active-noise conditions; horizontal lines denote the base-model and fully supervised references.}
\label{fig:noisy_label_results}
\end{figure*}

\subsubsection{Results and Discussion}
\label{sec:noisy_label_results}

Figure~\ref{fig:noisy_label_results} shows a consistent robustness benefit: OLR improves both ID and OOD averages at every tested noise ratio in both regimes. Under inactive noise, the largest gains are 8.8 ID points at $\rho=0.5$ and 5.4 OOD points at $\rho=0.9$. Under the more challenging active-noise regime, OLR improves ID accuracy by as much as 6.4 points at $\rho=0.5$ and OOD accuracy by as much as 8.1 points at $\rho=0.7$.

The matched comparisons are more informative than the absolute ranges: they show that refinement remains beneficial as both the amount and realizability of corrupted supervision change. At $\rho=0.9$ under active noise, both configurations deteriorate sharply and OLR provides only a limited recovery, indicating that refinement mitigates rather than eliminates extreme corruption. Across the remaining settings, however, the uniformly positive deltas demonstrate that DE-Venus can incorporate persistent label correction without changing the underlying GRPO execution path.

\subsection{Data Selection}
\label{sec:exp_data_selection}

The data-selection evaluation compares label-free strategies for two decisions: which examples should be annotated and which examples can be removed before RLVR training. All methods are implemented in DE-Venus and evaluated under both full-data and selected-subset training settings.

\subsubsection{Experimental Setup}
\label{sec:data_selection_setup}

\paragraph{Training and budget configuration.}
Models are trained on DAPO-Math-14K~\citep{yu2025dapo} with Qwen3-4B-Base as the backbone. Both selection settings annotate $\rho_a\approx29.3\%$ of the original examples. The first retains the full training set ($\rho_t=100\%$), whereas the second discards the bottom-ranked $42.1\%$ and trains on the remaining $\rho_t\approx57.9\%$.

\paragraph{Compared methods.}
Seven label-free selection strategies are considered: \textbf{Random}; \textbf{Consistency}~\citep{zuo2025ttrl}, which regards lower agreement among $G$ sampled answers as higher uncertainty; \textbf{Entropy}~\citep{huang2023look}, which uses the average output-token entropy; \textbf{Self-Certainty}~\citep{kang2025scalable}, which uses the average KL divergence between output-token distributions and a uniform distribution; \textbf{CoE}~\citep{wang2024latent}, which combines magnitude and angular variation of hidden states across layers; \textbf{CoT-Kinetics}~\citep{bi2025cot}, which measures semantic momentum and curvature energy of hidden states; and \textbf{PivotTrace}~\citep{zhu2026pivottrace}, which uses pivotal changes in long-range attention signals. Random ranks samples uniformly; the remaining non-random methods rank samples by descending estimated uncertainty. For every method, the top $|\mathcal{D}_a|$ examples are annotated and, in the subset-training setting, the bottom $|\mathcal{D}_d|$ examples are discarded.

\begin{figure*}[t]
\centering
\includegraphics[width=0.99\textwidth]{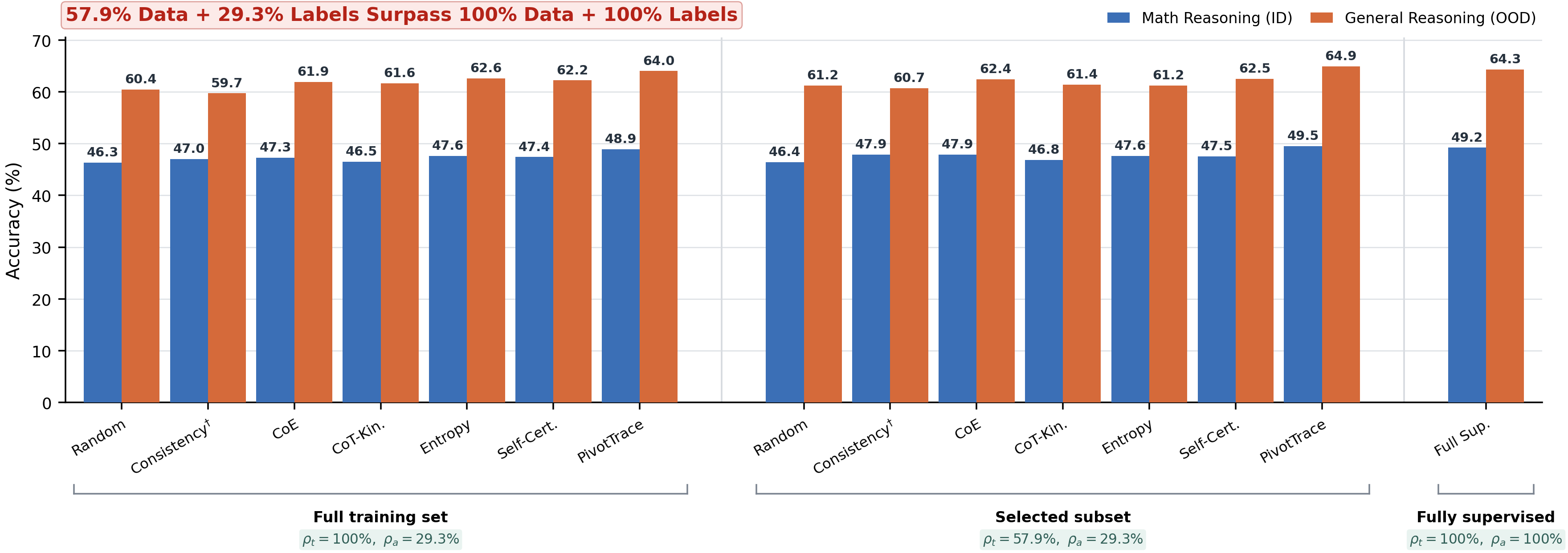}
\caption{\small ID and OOD average accuracy of Qwen3-4B-Base under different data-selection settings. Here, $\rho_t$ and $\rho_a$ denote the retained training-data and annotation fractions, respectively, while the fully supervised reference uses $\rho_t=\rho_a=100\%$. $\dagger$ marks methods requiring multiple stochastic inferences.}
\label{fig:select_results}
\end{figure*}

\subsubsection{Results and Discussion}
\label{sec:data_selection_results}

Figure~\ref{fig:select_results} shows that selection quality matters for both annotation and training efficiency. PivotTrace is strongest in both settings, reaching $48.9\%$ ID and $64.0\%$ OOD accuracy when the full training set is retained, and $49.5\%$ ID and $64.9\%$ OOD when training is restricted to the selected subset. Its improvement after discarding low-utility examples also shows that additional data is not uniformly beneficial.

The selected-subset result is the central efficiency outcome: using only $57.9\%$ of the training examples and $29.3\%$ of the annotations, PivotTrace exceeds the fully supervised full-data reference by 0.3 ID points and 0.6 OOD points. The remaining methods exhibit smaller or less consistent changes, which confirms that the gain cannot be attributed to reducing the dataset alone. DE-Venus therefore supports a joint reduction in training volume and annotation demand while retaining a common downstream RLVR pipeline.
\subsection{Business Scenario Validation}
\label{sec:business_scenario_validation}

The public studies above isolate individual framework capabilities under controlled conditions. We additionally evaluate whether those capabilities translate into operational gains in three business settings: loan-credit assignment with weak supervision, medical-empathy training with noisy rubric supervision, and intrinsic-safety training with data selection. Company names, internal project identifiers, model identifiers, and absolute business scores are suppressed; sample counts, selection ratios, public backbones, public benchmarks, and relative outcomes are retained. Results are normalized against the corresponding in-scenario baseline and should not be compared across scenarios.

Across the three settings, DE-Venus yields three concrete efficiency outcomes. With the same 500 labeled credit examples, reliability-aware use of 1,000 unlabeled examples improves the normalized business metric by up to \textbf{14 points}. In medical-empathy training, trajectory-based filtering removes \textbf{28\% of the training data} while remaining within 1.9 points of full-data training and 2.7 points above the untrained reference. In intrinsic-safety training, compact subsets preserve the principal safety metrics, improve selected capability metrics by up to \textbf{6.7\%}, and reduce the observed steps to convergence by \textbf{63--75\%}. These cases demonstrate that the framework can turn supervision quality into measurable annotation, computation, and iteration-cycle savings.

\subsubsection{Scenario A: Weakly Supervised Loan-Credit Assignment}
\label{sec:business_weak_supervision}

\paragraph{Scenario and setup.}
Scenario~A evaluates credit-limit interval prediction in an online-loan approval workflow characterized by a long-tailed customer distribution, limited annotation coverage, and ambiguous boundaries between adjacent intervals. The validation contains 500 labeled examples and 1,000 additional unlabeled examples. Labeled-only GRPO provides the reference. TTRL, EMRL, TraPO, and GeoMin retain the same 500-label budget and additionally use the unlabeled pool through majority-vote pseudo-labeling, entropy regularization, trajectory-based selection, and representation-distribution filtering, respectively.

\begin{figure*}[!t]
  \centering
  \includegraphics[width=0.96\textwidth]{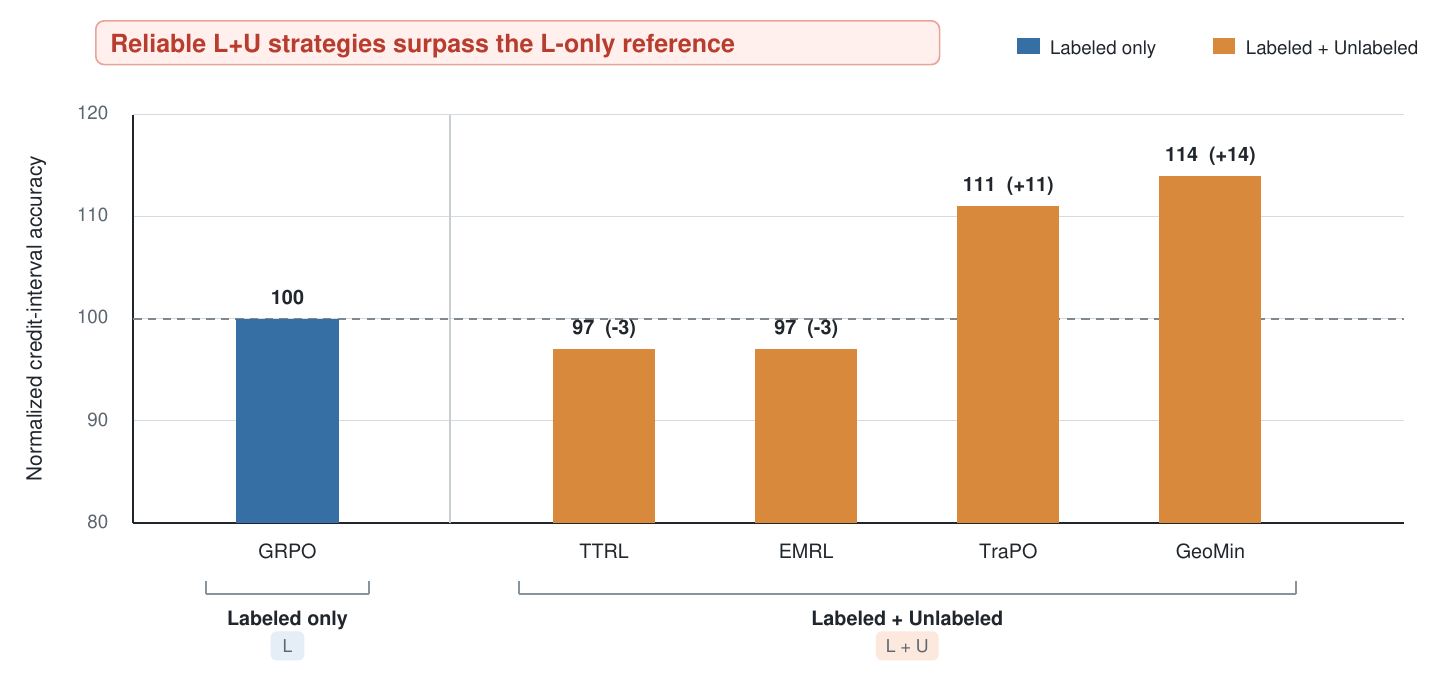}
  \caption{Normalized credit-interval accuracy in the loan-credit scenario. GRPO uses labeled data only (L), whereas TTRL, EMRL, TraPO, and GeoMin use the same labeled data plus an additional unlabeled set (L+U). Absolute business scores are suppressed and the labeled-only GRPO result is normalized to 100. Bar heights and direct labels report the rounded normalized index; parentheses give the change from the labeled-only reference. Higher values indicate better performance.}
  \label{fig:business_weak_supervision}
\end{figure*}

\paragraph{Results.}
Figure~\ref{fig:business_weak_supervision} normalizes the labeled-only GRPO result to 100. TraPO reaches 111 and GeoMin reaches 114, corresponding to gains of 11 and 14 points without requesting any additional labels. GeoMin provides the strongest result by filtering the unlabeled pool in representation space, while TraPO converts learning-trajectory evidence into a similarly substantial improvement.

The two direct baselines provide an important control: TTRL and EMRL both reach 97 despite receiving the same additional data. The benefit therefore does not arise from unlabeled volume alone. It appears only when DE-Venus identifies which weak signals are sufficiently reliable to influence the policy. In this scenario, reliability-aware selection turns a fixed annotation budget into a double-digit business gain, whereas indiscriminate use of the same unlabeled pool slightly degrades performance.

\subsubsection{Scenario B: Noisy Rubric Supervision for Medical Empathy}
\label{sec:business_noisy_supervision}

\paragraph{Scenario and setup.}
Scenario~B evaluates empathy training for a consumer-facing medical assistant, where responses must be professional, safe, understandable, and emotionally appropriate. Supervision is provided by rubric-based rewards, but some rubrics are ambiguous or weakly aligned with the desired behavior. DE-Venus uses the Qwen3-8B reward trajectory from epoch~0 to epoch~1 as an evidence signal. Among the 7,504 examples observed in both epochs, reward increases for 5,374, decreases for 2,047, and remains unchanged for 83. Filtering the decreasing and unchanged trajectories removes 2,130 low-confidence examples and retains approximately $72\%$ of the original data.

\begin{figure*}[!t]
  \centering
  \includegraphics[width=0.96\textwidth]{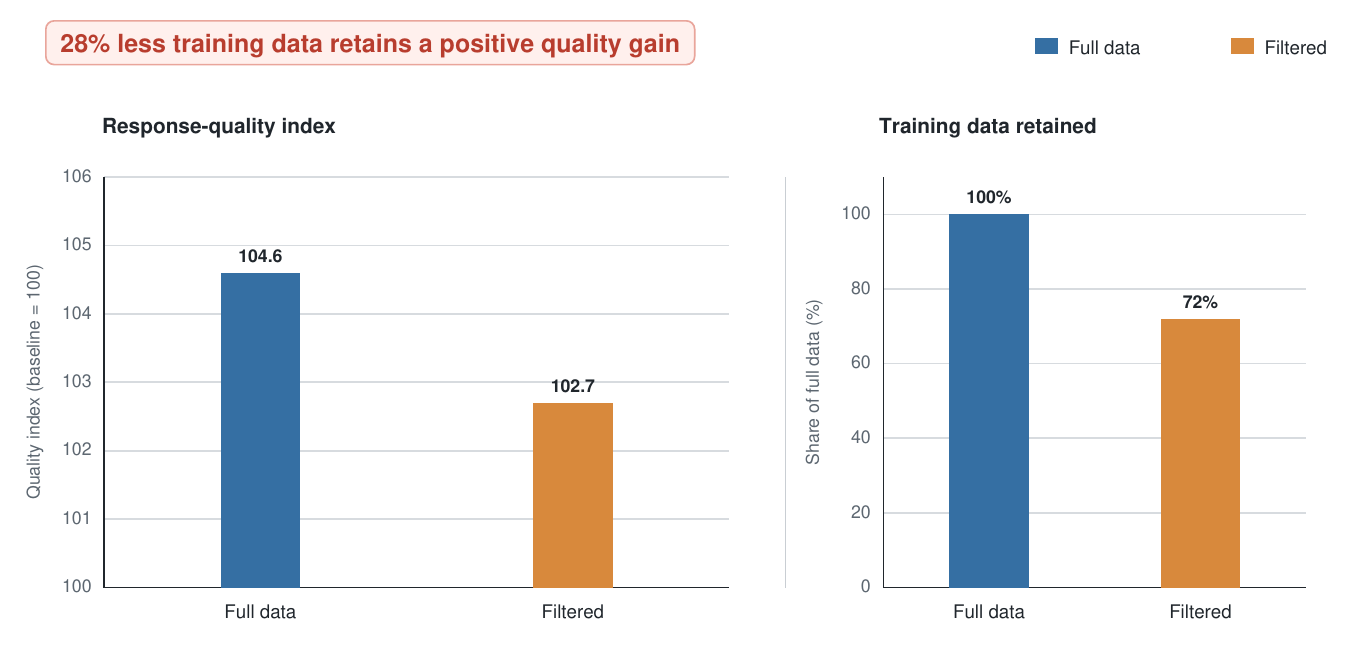}
  \caption{Medical-empathy results after trajectory-based rubric denoising. The left panel reports a response-quality index with the untrained reference normalized to 100 but not plotted as a separate bar; the right panel reports the fraction of training data retained. Absolute response-quality scores are suppressed, and higher normalized values indicate better performance.}
  \label{fig:business_noisy_supervision}
\end{figure*}

\paragraph{Results.}
Figure~\ref{fig:business_noisy_supervision} normalizes the untrained reference to 100. Full-data training reaches a response-quality index of 104.6, while the filtered configuration reaches 102.7 with only $72\%$ of the examples. Thus, DE-Venus removes more than one quarter of the training workload while remaining within 1.9 points of full-data training and preserving a 2.7-point gain over the untrained model. The objective in this scenario is cost-controlled denoising rather than a higher peak score: reward trajectories provide a reusable confidence signal that reduces training volume without additional annotation and with a bounded quality trade-off.

\subsubsection{Scenario C: Data Selection for Intrinsic Safety}
\label{sec:business_data_selection}

\begin{figure*}[!t]
\centering
\includegraphics[width=0.99\textwidth]{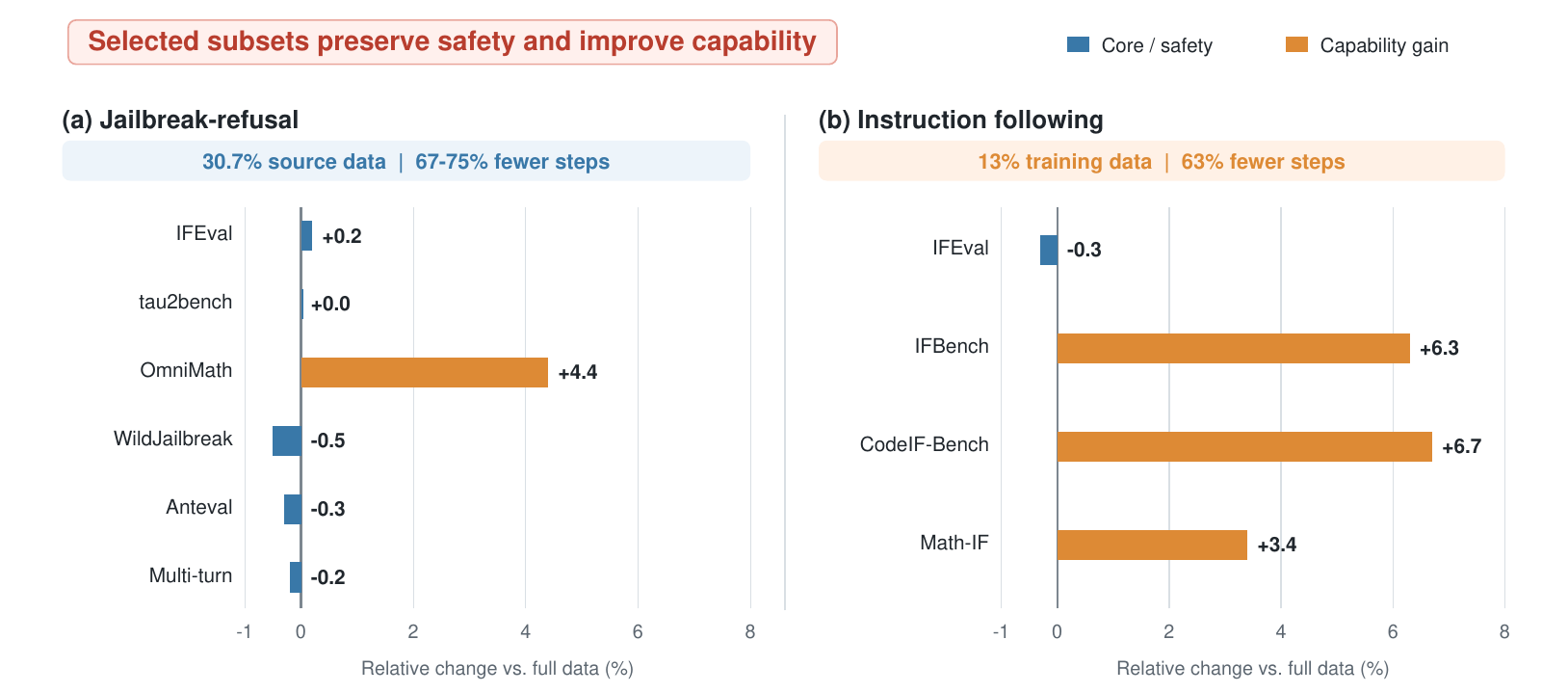}
\caption{Relative capability changes at the selected operating points in Scenario~C. Panel~(a) retains $30.67\%$ of the mathematics and instruction source pool for jailbreak-refusal training, while Panel~(b) retains $13\%$ of the instruction-following data. Bars report changes against the corresponding full-data baseline, and the observed reductions in convergence steps are shown above each panel.}
\label{fig:business_intrinsic_safety}
\end{figure*}

\paragraph{Jailbreak-refusal task.}
The first intrinsic-safety validation selects the strongly verifiable RLVR portion of a jailbreak-refusal training mixture while holding the RL hyperparameters, reward function, and execution environment fixed. From a 3,225-example mathematics and instruction source pool, DE-Venus retains 989 examples, or $30.67\%$ of that pool; its share of the complete training mixture falls from $11.46\%$ to $3.82\%$. As shown in Figure~\ref{fig:business_intrinsic_safety}(a), the selected-data run remains within $0.5\%$ of the full-data baseline on IFEval, tau2bench, WildJailbreak, Anteval, and the stepwise multi-turn attack evaluation, while improving OmniMath by $4.4\%$. It converges in approximately 200 steps instead of 600--800, a reduction of roughly $67$--$75\%$ in observed optimization steps.

\paragraph{Instruction-following task.}
The second validation starts from 2,380 RL examples and constructs subsets retaining $13\%$, $23\%$, and $38\%$ of the full data. Figure~\ref{fig:business_intrinsic_safety}(b) shows that the smallest, $13\%$ subset offers the strongest operating point: it approximately matches the full-data reference on IFEval ($-0.3\%$) while improving IFBench by $6.3\%$, CodeIF-Bench by $6.7\%$, and Math-IF by $3.4\%$. This run converges in approximately 110 steps rather than 300, reducing the observed training cycle by about $63\%$. The higher-retention subsets produce smaller and mixed changes, confirming that the value comes from identifying high-utility examples rather than simply retaining more data.

Across both intrinsic-safety tasks, data selection functions as a resource-allocation mechanism: compact subsets preserve the full-data capability level on sensitive safety metrics, improve selected general capabilities, and shorten the iteration cycle substantially.

%% file: sections/conclusion.tex
\section{Conclusion}

We present DE-Venus, a unified framework for data-efficient reinforcement learning with verifiable rewards. DE-Venus organizes methods around a common supervision lifecycle: selecting useful training or annotation candidates before optimization, constructing learning signals from partially labeled or unlabeled data, and refining unreliable supervision during training. By implementing these interventions as a lightweight layer over verl, the framework preserves a shared distributed RL substrate while localizing method-specific logic to explicit data, reward, filtering, relabeling, and advantage-transformation interfaces. This design enables heterogeneous data-efficient RLVR methods to be implemented, evaluated, and extended without repeatedly rebuilding the underlying rollout and optimization system.

Our experiments evaluate DE-Venus across weakly supervised learning, noisy-label learning, and data selection, as well as representative business scenarios. The results demonstrate that carefully constructed weak supervision can approach or exceed fully supervised training with substantially fewer verified labels, that Online Label Refinement improves robustness across a wide range of label-noise conditions, and that model-derived selection signals can reduce annotation and training requirements while maintaining competitive performance. Together, these results show that data efficiency in RLVR is best treated as an end-to-end supervision problem rather than as an isolated choice of dataset size or reward function. We hope DE-Venus provides a practical and reproducible foundation for developing data-efficient reasoning algorithms and for extending reliable weak-supervision techniques to broader models, tasks, and domains.

%% file: main.bib
@article{guo2025deepseek,
  title={Deepseek-r1: Incentivizing reasoning capability in llms via reinforcement learning},
  author={Guo, Daya and Yang, Dejian and Zhang, Haowei and Song, Junxiao and Zhang, Ruoyu and Xu, Runxin and Zhu, Qihao and Ma, Shirong and Wang, Peiyi and Bi, Xiao and others},
  journal={arXiv preprint arXiv:2501.12948},
  year={2025}
}

@article{jaech2024openai,
  title={Openai o1 system card},
  author={Jaech, Aaron and Kalai, Adam and Lerer, Adam and Richardson, Adam and El-Kishky, Ahmed and Low, Aiden and Helyar, Alec and Madry, Aleksander and Beutel, Alex and Carney, Alex and others},
  journal={arXiv preprint arXiv:2412.16720},
  year={2024}
}

@article{team2025kimi,
  title={Kimi k1. 5: Scaling reinforcement learning with llms},
  author={Team, Kimi and Du, Angang and Gao, Bofei and Xing, Bowei and Jiang, Changjiu and Chen, Cheng and Li, Cheng and Xiao, Chenjun and Du, Chenzhuang and Liao, Chonghua and others},
  journal={arXiv preprint arXiv:2501.12599},
  year={2025}
}

@inproceedings{ahmadian2024back,
  title={Back to Basics: Revisiting {REINFORCE}-Style Optimization for Learning from Human Feedback in {LLM}s},
  author={Ahmadian, Arash and Cremer, Chris and Gall{\'e}, Matthias and Fadaee, Marzieh and Kreutzer, Julia and Pietquin, Olivier and {\"U}st{\"u}n, Ahmet and Hooker, Sara},
  booktitle={Proceedings of the 62nd Annual Meeting of the Association for Computational Linguistics (Volume 1: Long Papers)},
  pages={12248--12267},
  year={2024},
  publisher={Association for Computational Linguistics},
  doi={10.18653/v1/2024.acl-long.662}
}

@article{hu2025reinforcepp,
  title={{REINFORCE++}: Stabilizing Critic-Free Policy Optimization with Global Advantage Normalization},
  author={Hu, Jian and Liu, Jason Klein and Xu, Haotian and Shen, Wei},
  journal={arXiv preprint arXiv:2501.03262},
  year={2025}
}

@misc{grpo,
      title={DeepSeekMath: Pushing the Limits of Mathematical Reasoning in Open Language Models}, 
      author={Zhihong Shao and Peiyi Wang and Qihao Zhu and Runxin Xu and Junxiao Song and Xiao Bi and Haowei Zhang and Mingchuan Zhang and Y. K. Li and Y. Wu and Daya Guo},
      year={2024},
      eprint={2402.03300},
      archivePrefix={arXiv},
      primaryClass={cs.CL},
      url={https://arxiv.org/abs/2402.03300}, 
}

@article{drgrpo,
  title={Understanding R1-Zero-Like Training: A Critical Perspective},
  author={Zichen Liu and Changyu Chen and Wenjun Li and Penghui Qi and Tianyu Pang and Chao Du and Wee Sun Lee and Min Lin},
  journal={arXiv preprint arXiv:2503.20783},
  year={2025}
}

@misc{orz,
      title={Open-Reasoner-Zero: An Open Source Approach to Scaling Up Reinforcement Learning on the Base Model}, 
      author={Jingcheng Hu and Yinmin Zhang and Qi Han and Daxin Jiang and Xiangyu Zhang and Heung-Yeung Shum},
      year={2025},
      eprint={2503.24290},
      archivePrefix={arXiv},
      primaryClass={cs.LG},
      url={https://arxiv.org/abs/2503.24290}, 
}

@misc{li2024numinamath,
  author       = {Jia Li and Edward Beeching and Lewis Tunstall and Ben Lipkin and Roman Soletskyi and Shengyi Huang and Kashif Rasul and Longhui Yu and Albert Q. Jiang and Ziju Shen and others},
  title        = {Numinamath: The largest public dataset in AI4Maths with 860k pairs of competition math problems and solutions},
  year         = {2024},
  howpublished = {\url{https://huggingface.co/datasets/Numinamath}},
  note         = {Hugging Face repository, 13:9}
}

@inproceedings{he2024olympiadbench,
  title={Olympiadbench: A challenging benchmark for promoting agi with olympiad-level bilingual multimodal scientific problems},
  author={He, Chaoqun and Luo, Renjie and Bai, Yuzhuo and Hu, Shengding and Thai, Zhen and Shen, Junhao and Hu, Jinyi and Han, Xu and Huang, Yujie and Zhang, Yuxiang and others},
  booktitle={Proceedings of the 62nd Annual Meeting of the Association for Computational Linguistics (Volume 1: Long Papers)},
  pages={3828--3850},
  year={2024}
}

@article{lewkowycz2022solving,
  title={Solving quantitative reasoning problems with language models},
  author={Lewkowycz, Aitor and Andreassen, Anders and Dohan, David and Dyer, Ethan and Michalewski, Henryk and Ramasesh, Vinay and Slone, Ambrose and Anil, Cem and Schlag, Imanol and Gutman-Solo, Theo and others},
  journal={Advances in neural information processing systems},
  volume={35},
  pages={3843--3857},
  year={2022}
}

@article{hendrycks2021measuring,
  title={Measuring mathematical problem solving with the math dataset},
  author={Hendrycks, Dan and Burns, Collin and Kadavath, Saurav and Arora, Akul and Basart, Steven and Tang, Eric and Song, Dawn and Steinhardt, Jacob},
  journal={arXiv preprint arXiv:2103.03874},
  year={2021}
}

@article{clark2018think,
  title={Think you have solved question answering? try arc, the ai2 reasoning challenge},
  author={Clark, Peter and Cowhey, Isaac and Etzioni, Oren and Khot, Tushar and Sabharwal, Ashish and Schoenick, Carissa and Tafjord, Oyvind},
  journal={arXiv preprint arXiv:1803.05457},
  year={2018}
}

@inproceedings{rein2024gpqa,
  title={Gpqa: A graduate-level google-proof q\&a benchmark},
  author={Rein, David and Hou, Betty Li and Stickland, Asa Cooper and Petty, Jackson and Pang, Richard Yuanzhe and Dirani, Julien and Michael, Julian and Bowman, Samuel R},
  booktitle={First Conference on Language Modeling},
  year={2024}
}

@article{wang2024mmlu,
  title={Mmlu-pro: A more robust and challenging multi-task language understanding benchmark},
  author={Wang, Yubo and Ma, Xueguang and Zhang, Ge and Ni, Yuansheng and Chandra, Abhranil and Guo, Shiguang and Ren, Weiming and Arulraj, Aaran and He, Xuan and Jiang, Ziyan and others},
  journal={Advances in Neural Information Processing Systems},
  volume={37},
  pages={95266--95290},
  year={2024}
}

@article{zuo2025ttrl,
  title={TTRL: Test-Time Reinforcement Learning},
  author={Zuo, Yuxin and Zhang, Kaiyan and Qu, Shang and Sheng, Li and Zhu, Xuekai and Qi, Biqing and Sun, Youbang and Cui, Ganqu and Ding, Ning and Zhou, Bowen},
  journal={arXiv preprint arXiv:2504.16084},
  year={2025}
}

@article{zhao2025absolutezero,
    title={Absolute Zero: Reinforced Self-play Reasoning with Zero Data}, 
    author={Andrew Zhao and Yiran Wu and Yang Yue and Tong Wu and Quentin Xu and Yang Yue and Matthieu Lin and Shenzhi Wang and Qingyun Wu and Zilong Zheng and Gao Huang},
    journal={arXiv preprint arXiv:2505.03335},
    year={2025}
}

@article{yu2025dapo,
  title={Dapo: An open-source llm reinforcement learning system at scale},
  author={Yu, Qiying and Zhang, Zheng and Zhu, Ruofei and Yuan, Yufeng and Zuo, Xiaochen and Yue, Yu and Dai, Weinan and Fan, Tiantian and Liu, Gaohong and Liu, Lingjun and others},
  journal={arXiv preprint arXiv:2503.14476},
  year={2025}
}

@article{zheng2025group,
  title={Group sequence policy optimization},
  author={Zheng, Chujie and Liu, Shixuan and Li, Mingze and Chen, Xiong-Hui and Yu, Bowen and Gao, Chang and Dang, Kai and Liu, Yuqiong and Men, Rui and Yang, An and others},
  journal={arXiv preprint arXiv:2507.18071},
  year={2025}
}

@article{zhao2025learning,
  title={Learning to reason without external rewards},
  author={Zhao, Xuandong and Kang, Zhewei and Feng, Aosong and Levine, Sergey and Song, Dawn},
  journal={arXiv preprint arXiv:2505.19590},
  year={2025}
}

@article{agarwal2025unreasonable,
  title={The unreasonable effectiveness of entropy minimization in llm reasoning},
  author={Agarwal, Shivam and Zhang, Zimin and Yuan, Lifan and Han, Jiawei and Peng, Hao},
  journal={arXiv preprint arXiv:2505.15134},
  year={2025}
}

@article{zhang2025co,
  title={Co-Reward: Self-supervised Reinforcement Learning for Large Language Model Reasoning via Contrastive Agreement},
  author={Zhang, Zizhuo and Zhu, Jianing and Ge, Xinmu and Zhao, Zihua and Zhou, Zhanke and Li, Xuan and Feng, Xiao and Yao, Jiangchao and Han, Bo},
  journal={arXiv preprint arXiv:2508.00410},
  year={2025}
}

@article{li2025limr,
  title={Limr: Less is more for rl scaling},
  author={Li, Xuefeng and Zou, Haoyang and Liu, Pengfei},
  journal={arXiv preprint arXiv:2502.11886},
  year={2025}
}

@article{bi2025cot,
  title={Cot-kinetics: A theoretical modeling assessing lrm reasoning process},
  author={Bi, Jinhe and Yan, Danqi and Wang, Yifan and Huang, Wenke and Chen, Haokun and Wan, Guancheng and Ye, Mang and Xiao, Xun and Schuetze, Hinrich and Tresp, Volker and others},
  journal={arXiv preprint arXiv:2505.13408},
  year={2025}
}

@article{wang2024latent,
  title={Latent space chain-of-embedding enables output-free llm self-evaluation},
  author={Wang, Yiming and Zhang, Pei and Yang, Baosong and Wong, Derek F and Wang, Rui},
  journal={arXiv preprint arXiv:2410.13640},
  year={2024}
}

@article{he2025deepmath,
  title={Deepmath-103k: A large-scale, challenging, decontaminated, and verifiable mathematical dataset for advancing reasoning},
  author={He, Zhiwei and Liang, Tian and Xu, Jiahao and Liu, Qiuzhi and Chen, Xingyu and Wang, Yue and Song, Linfeng and Yu, Dian and Liang, Zhenwen and Wang, Wenxuan and others},
  journal={arXiv preprint arXiv:2504.11456},
  year={2025}
}

@article{yan2025verifybench,
  title={Verifybench: Benchmarking reference-based reward systems for large language models},
  author={Yan, Yuchen and Jiang, Jin and Ren, Zhenbang and Li, Yijun and Cai, Xudong and Liu, Yang and Xu, Xin and Zhang, Mengdi and Shao, Jian and Shen, Yongliang and others},
  journal={arXiv preprint arXiv:2505.15801},
  year={2025}
}

@inproceedings{tang2025towards,
  title={Towards Transferable Personality Representation Learning based on Triplet Comparisons and Its Applications},
  author={Tang, Kai and Wang, Rui and Zhu, Renyu and Lin, Minmin and Ding, Xiao and Lv, Tangjie and Fan, Changjie and Wu, Runze and Wang, Haobo},
  booktitle={Proceedings of the 2025 Conference on Empirical Methods in Natural Language Processing},
  pages={10061--10077},
  year={2025}
}

@article{yang2025qwen3,
  title={Qwen3 technical report},
  author={Yang, An and Li, Anfeng and Yang, Baosong and Zhang, Beichen and Hui, Binyuan and Zheng, Bo and Yu, Bowen and Gao, Chang and Huang, Chengen and Lv, Chenxu and others},
  journal={arXiv preprint arXiv:2505.09388},
  year={2025}
}

@inproceedings{sheng2024hybridflow_verl,
  title     = {HybridFlow: A Flexible and Efficient RLHF Framework},
  author    = {Sheng, Guangming and Zhang, Chi and Ye, Zilingfeng and Wu, Xibin and Zhang, Wang and Zhang, Ru and Peng, Yanghua and Lin, Haibin and Wu, Chuan},
  booktitle = {Proceedings of the Twentieth European Conference on Computer Systems},
  pages     = {1279--1297},
  year      = {2025},
  doi       = {10.1145/3689031.3696075}
}

@inproceedings{bae2025online,
  title={Online difficulty filtering for reasoning oriented reinforcement learning},
  author={Bae, Sanghwan and Hong, Jiwoo and Lee, Min Young and Kim, Hanbyul and Nam, JeongYeon and Kwak, Donghyun},
  booktitle={Proceedings of the 19th Conference of the European Chapter of the Association for Computational Linguistics (Volume 1: Long Papers)},
  pages={700--719},
  year={2026}
}

@article{zeng2025cures,
  title={CurES: From Gradient Analysis to Efficient Curriculum Learning for Reasoning LLMs},
  author={Zeng, Yongcheng and Sun, Zexu and Ji, Bokai and Min, Erxue and Cai, Hengyi and Wang, Shuaiqiang and Yin, Dawei and Zhang, Haifeng and Chen, Xu and Wang, Jun},
  journal={arXiv preprint arXiv:2510.01037},
  year={2025}
}

@article{huang2023uncertainty,
  title={Look before you leap: An exploratory study of uncertainty measurement for large language models},
  author={Huang, Yuheng and Song, Jiayang and Wang, Zhijie and Zhao, Shengming and Chen, Huaming and Juefei-Xu, Felix and Ma, Lei},
  journal={arXiv preprint arXiv:2307.10236},
  year={2023}
}

@article{kang2026selfcertainty,
  title={Scalable best-of-n selection for large language models via self-certainty},
  author={Kang, Zhewei and Zhao, Xuandong and Song, Dawn},
  journal={Advances in neural information processing systems},
  volume={38},
  pages={19720--19745},
  year={2026}
}

@article{zhu2026pivottrace,
  title={Smart Picks in the Dark: Towards Efficient RLVR for Reasoning via Tracing Metacognitive Pivots},
  author={Zhu, Guangcheng and Yang, Shenzhi and Wang, Haobo and Zheng, Xing and MA, Yingfan and Feng, Xuening and Chen, Zhongqi and Song, Bowen and Wang, Weiqiang and Chen, Gang},
  journal={arXiv preprint arXiv:2606.04503},
  year={2026}
}

@article{zhang2025corewarding,
  title={Co-rewarding: Stable Self-supervised RL for Eliciting Reasoning in Large Language Models},
  author={Zhang, Zizhuo and Zhu, Jianing and Ge, Xinmu and Zhao, Zihua and Zhou, Zhanke and Li, Xuan and Feng, Xiao and Yao, Jiangchao and Han, Bo},
  journal={arXiv preprint arXiv:2508.00410},
  year={2025}
}

@article{yang2025trapo,
  title={TraPO: A Semi-Supervised Reinforcement Learning Framework for Boosting LLM Reasoning},
  author={Yang, Shenzhi and Zhu, Guangcheng and Zheng, Xing and MA, Yingfan and Chen, Zhongqi and Song, Bowen and Wang, Weiqiang and Zhao, Junbo and Chen, Gang and Wang, Haobo},
  journal={arXiv preprint arXiv:2512.13106},
  year={2025}
}

@article{zhu2026geomin,
  title={GeoMin: Data-Efficient Semi-Supervised RLVR via Geometric Distribution Modeling},
  author={Zhu, Guangcheng and Yang, Shenzhi and Wang, Haobo and Zheng, Xing and MA, Yingfan and Feng, Xuening and Chen, Zhongqi and Tang, Kai and Zang, Zhengqing and Song, Bowen and others},
  journal={arXiv preprint arXiv:2606.04516},
  year={2026}
}

@article{yang2026can,
  title={Can LLMs Learn to Reason Robustly under Noisy Supervision?},
  author={Yang, Shenzhi and Zhu, Guangcheng and Song, Bowen and Li, Sharon and Wang, Haobo and Zheng, Xing and Ma, Yingfan and Chen, Zhongqi and Wang, Weiqiang and Chen, Gang},
  journal={arXiv preprint arXiv:2604.03993},
  year={2026}
}

@article{huang2023look,
  title={Look before you leap: An exploratory study of uncertainty measurement for large language models},
  author={Huang, Yuheng and Song, Jiayang and Wang, Zhijie and Zhao, Shengming and Chen, Huaming and Juefei-Xu, Felix and Ma, Lei},
  journal={arXiv preprint arXiv:2307.10236},
  year={2023}
}

@article{kang2025scalable,
  title={Scalable best-of-n selection for large language models via self-certainty},
  author={Kang, Zhewei and Zhao, Xuandong and Song, Dawn},
  journal={arXiv preprint arXiv:2502.18581},
  year={2025}
}
